\documentclass{article} 
\usepackage{iclr2026_conference,times}

\usepackage{amsmath,amsfonts,bm}

\def\eqref#1{equation~\ref{#1}}

\def\1{\bm{1}}

\DeclareMathAlphabet{\mathsfit}{\encodingdefault}{\sfdefault}{m}{sl}
\SetMathAlphabet{\mathsfit}{bold}{\encodingdefault}{\sfdefault}{bx}{n}

\usepackage[colorlinks, citecolor=teal, urlcolor=blue, linkcolor=black]{hyperref}

\usepackage{url}

\usepackage[ruled,vlined]{algorithm2e}
\usepackage[textsize=tiny]{todonotes}
\usepackage[export]{adjustbox}
\usepackage[most,skins]{tcolorbox}
\usepackage[dvipsnames]{xcolor}
\usepackage{colortbl}
\usepackage{amsmath}
\usepackage{mathtools}
\usepackage{tikz-cd} 
\usepackage{subfigure}    
\usepackage{mdframed}
\usepackage{pifont}
\usepackage{tabularx} 
\usepackage{threeparttable}
\usepackage{subcaption}
\usepackage{array}
\usepackage{svg}
\usepackage{multirow}
\usepackage{minted}
\usepackage{mathrsfs} 
\usepackage{booktabs}
\usepackage{enumitem}
\usepackage{newfloat}
\usepackage{listings}
\usepackage{setspace}
\usepackage{colortbl}
\usepackage{arydshln}
\usepackage{amsfonts}
\usepackage{makecell}
\usepackage{caption}
\usepackage{cleveref}
\usepackage{wrapfig}
\usepackage{amssymb}
\usepackage{bbding}
\usepackage{pifont}
\usepackage{xspace}
\usepackage{float}
\usepackage{bbm}
\usepackage{bm}

\usepackage[dvipsnames]{xcolor}
\usepackage{tikz}
\usetikzlibrary{backgrounds}
\usetikzlibrary{arrows,shapes}
\usetikzlibrary{tikzmark}
\usetikzlibrary{calc}

\usepackage{arydshln} 
\definecolor{rxkdarkdarkblue}{RGB}{95, 108, 161} 
\definecolor{rxkdarkblue}{RGB}{213, 218, 233} 
\definecolor{rxklightblue}{RGB}{242, 244, 250} 

\newcolumntype{x}[1]{>{\centering\arraybackslash}p{#1pt}}
\newcolumntype{I}{!{\vrule width 1pt}}
\newcommand{\myhyperlink}[3][black]{\hyperlink{#2}{\color{#1}{#3}}}

\makeatletter
\newcommand{\thickhline}{%
    \noalign {\ifnum 0=`}\fi \hrule height 1pt
    \futurelet \reserved@a \@xhline
}
\makeatother

\makeatletter
\newenvironment{fullitemize}{
\vspace{-5pt}
\begin{itemize}[leftmargin=*]
\setlength{\itemsep}{5pt}
\setlength{\parsep}{-5pt}
\setlength{\parskip}{-3pt}
\setlength{\leftmargin}{-10pt}}
{\end{itemize}
\vspace{-5pt}
}
\makeatother

\tcbset{highlight math/.append style={left=0mm,right=0mm,top=0mm,bottom=0mm, colframe=white}}

\makeatletter
\DeclareRobustCommand\onedot{\futurelet\@let@token\@onedot}

\def\@onedot{\ifx\@let@token.\else.\null\fi\xspace}
\def\eg{\textit{e.g}\onedot} 
\def\ie{\textit{i.e}\onedot}

\newcommand{\pub}[1]{{\color{gray}{\footnotesize{[{#1}]}}}}

\definecolor{mydarkdarkred}{RGB}{182, 58, 43} 
\definecolor{mydarkdarkred2}{RGB}{180, 92, 67} 
\definecolor{mydarkred}{RGB}{245, 220, 215}  
\definecolor{mylightred}{RGB}{251, 244, 242}  

\definecolor{mydarkdarkblue}{RGB}{17, 24, 129} 
\definecolor{mydarkdarkblue2}{RGB}{92, 108, 165} 
\definecolor{mydarkblue}{RGB}{211, 218, 234} 
\definecolor{mylightblue}{RGB}{241, 244, 250}

\definecolor{DarkBlue}{RGB}{64,101,149}

\definecolor{mydarkdarkgreen}{RGB}{93, 150, 74} 
\definecolor{mydarkgreen}{RGB}{216, 233, 199}  
\definecolor{mylightgreen}{RGB}{245, 249, 241}

\definecolor{darkred}{RGB}{139,0,0}  

\definecolor{DarkGreen}{RGB}{42,110,63}
\definecolor{DarkYellow}{RGB}{160, 115, 0}

\definecolor{mydarkyellow}{RGB}{216, 214, 196}   
\definecolor{mymiddleyellow}{RGB}{229, 228, 218}
\definecolor{mylightyellow}{RGB}{245, 245, 240} 

\crefname{proposition}{Prop.}{Props.}
\crefname{section}{Sec.}{Secs.}
\crefname{table}{Tab.}{Tabs.}

\title{MAPO: Mixed Advantage Policy Optimization}

\author{Wenke Huang$^{1}$ \; Quan Zhang$^{2\dagger}$ \; \\
\textbf{Yiyang Fang}$^{1}$ \; \textbf{Jian Liang}$^{1}$ \; \textbf{Xuankun Rong}$^{1}$ \;  \textbf{Huanjin Yao}$^{4}$ \; \\
\textbf{Guancheng Wan}$^{1}$ \; \textbf{Ke Liang}$^{3}$ \;  \textbf{Wenwen He}$^{1}$ \; \\
\textbf{Mingjun Li}$^{2\dagger}$ \; 
\textbf{Leszek Rutkowski}$^{5}$ \; 
\textbf{Mang Ye}$^{1}$\thanks{Corresponding author. $\dagger$ Project Leader. Work done by Wenke Huang during internship at ByteDance.}\; \quad 
\textbf{Bo Du}$^{1*}$ \;  \textbf{Dacheng Tao}$^{4}$\\
$^{1}$ Wuhan University \;
$^{2}$ ByteDance \;
$^{3}$ National University of Defense Technology \\
$^{4}$ Nanyang Technological University \;
$^{5}$ The AGH University of Krakow
}

\iclrfinalcopy 

\begin{document}

\maketitle

\begin{abstract}
Recent advances in reinforcement learning for foundation model post-training, such as Group Relative Policy Optimization (GRPO), have significantly improved the performance of foundation models on reasoning tasks. Notably, the advantage function serves as a central mechanism for ranking the importance of trajectory candidates. However, existing exploration methods encounter both advantage reversion and advantage mirror problems, which hinder the reasonable allocation of advantage across different query samples. In this work, we propose an easy but effective GRPO strategy, \textbf{M}ixed \textbf{A}dvantage \textbf{P}olicy \textbf{O}ptimization (\textbf{MAPO}). We reveal the trajectory certainty characteristic and propose the use of advantage percent deviation for high-certainty trajectories. Furthermore, we dynamically reweight the advantage function for samples with varying trajectory certainty, thereby adaptively configuring the advantage function to account for sample-specific characteristics. Ablation studies with different advantage variants, along with empirical evaluations on existing GRPO explorations, confirm our effectiveness, showcasing its ability to mitigate prior advantage function shortcomings. As a result, foundation models trained with our approach produce more stable and accurate reasoning performance across diverse tasks.
\end{abstract}

\newcommand{\attention}{{Attention}}
\newcommand{\bert}{{BERT}}
\newcommand{\vit}{{ViT}}
\newcommand{\aft}{{AFT}}
\newcommand{\cpvt}{{CPVT}}
\newcommand{\alphaedit}{{AlphaEdit}}
\newcommand{\deit}{{DeiT}}
\newcommand{\convit}{{ConViT}}
\newcommand{\orpo}{{ORPO}}
\newcommand{\remedy}{{REMEDY}}

\newcommand{\clora}{{CLoRA}} 
\newcommand{\ceit}{{CeiT}}
\newcommand{\greedyprune}{{GreedyPrune}}
\newcommand{\loramoe}{{LoRAMoE}}
\newcommand{\ronecompress}{{R1Compress}}
\newcommand{\visonpert}{{VP}}
\newcommand{\omnidpo}{{Omni-DPO}}
\newcommand{\treerpo}{{TreeRPO}}
\newcommand{\srft}{{SRFT}}
\newcommand{\drgrpo}{{Dr. GRPO}}
\newcommand{\palm}{{PaLM}}
\newcommand{\rlif}{{RLIF}}
\newcommand{\seedgrpo}{{SEED-GRPO}}
\newcommand{\groudrone}{{Ground-R1}}
\newcommand{\vlrethinker}{{VL-Rethinker}}
\newcommand{\cogvlm}{{CogVLM}}
\newcommand{\grpocare}{{GRPO-CARE}}
\newcommand{\wdone}{{wd1}}
\newcommand{\krpo}{{KRPO}}
\newcommand{\mgrpo}{{MGRPO}}
\newcommand{\sia}{{SIA}}
\newcommand{\sgrpo}{{S-GRPO}}
\newcommand{\vstar}{{V$^*$}}
\newcommand{\grpolead}{{GRPO-LEAD}}
\newcommand{\grpolambda}{{GRPO-$\lambda$}}
\newcommand{\visionrone}{{Vision-R1}}
\newcommand{\hintgrpo}{Hint-GRPO}
\newcommand{\asrr}{{ASRR}}
\newcommand{\sua}{{SUA}}
\newcommand{\opcm}{{OPCM}}
\newcommand{\dapo}{{DAPO}}
\newcommand{\gpg}{{GPG}}
\newcommand{\dsf}{{DSF}}
\newcommand{\visionreasoner}{{VisionReasoner}}
\newcommand{\dids}{{DIDS}}
\newcommand{\sophicvlrone}{{SophiaVL-R1}}
\newcommand{\noisyrollout}{{NoisyRollout}}
\newcommand{\pure}{{PURE}}
\newcommand{\ttrl}{{TTRL}}
\newcommand{\oft}{{OFT}}
\newcommand{\hidellava}{{HiDe-LLaVA}}
\newcommand{\sewa}{{SeWA}}
\newcommand{\nft}{{NFT}}
\newcommand{\saft}{{SAFT}}
\newcommand{\badtoken}{{BadToken}}
\newcommand{\octopus}{{Octopus}}
\newcommand{\sharegrpo}{Share-GRPO}
\newcommand{\damo}{{DAMO}}
\newcommand{\ctrap}{{CTRAP}}
\newcommand{\panacea}{{Panacea}}
\newcommand{\memvr}{{MemVR}}
\newcommand{\localvit}{{LocalViT}}
\newcommand{\halva}{{HALVA}}
\newcommand{\swin}{{Swin}}
\newcommand{\sefe}{{SEFE}}
\newcommand{\pissa}{{PiSSA}}
\newcommand{\simignore}{{Simignore}}
\newcommand{\dola}{{DOLA}}
\newcommand{\cosacl}{{CoS}}
\newcommand{\adastar}{{AdaSTaR}}
\newcommand{\olora}{{O-LoRA}}
\newcommand{\erec}{{EReC}}
\newcommand{\hira}{{HiRA}}
\newcommand{\prolora}{{PRoLoRA}}
\newcommand{\mixlora}{{MixLoRA}}
\newcommand{\deco}{{DeCo}}
\newcommand{\flipattack}{{FLIPATTACK}}
\newcommand{\var}{{VAR}}
\newcommand{\dpc}{{DPC}}
\newcommand{\autoprompt}{{AutoPrompt}}
\newcommand{\maba}{{MABA}}
\newcommand{\lpaqa}{{LPAQA}}
\newcommand{\guardreasoner}{{GuardReasoner}}
\newcommand{\vrft}{{Visual-RFT}}
\newcommand{\bathe}{{BaThe}}
\newcommand{\sat}{{SAT}}
\newcommand{\milora}{{MiLoRA}}
\newcommand{\atmmerge}{{ATM}}
\newcommand{\opera}{{OPERA}}
\newcommand{\cvt}{{CvT}}
\newcommand{\eppo}{{EPPO}}
\newcommand{\fsd}{{FSD}}
\newcommand{\nlora}{{NLoRA}}
\newcommand{\adagc}{{AdaGC}}
\newcommand{\vhm}{{VHM}}
\newcommand{\cofitune}{{CoFiTune}}
\newcommand{\child}{{CHILD}}
\newcommand{\loraga}{{LoRA-GA}}
\newcommand{\galore}{{GaLore}}
\newcommand{\id}{{ID}}
\newcommand{\lors}{{LoRS}}
\newcommand{\vacode}{{VaCoDe}}
\newcommand{\arf}{{ARF}}
\newcommand{\ecso}{{ECSO}}
\newcommand{\lorarar}{{LoRA.rar}}
\newcommand{\janus}{{Janus}}
\newcommand{\cart}{{CART}}
\newcommand{\fastvlm}{{FastVLM}}
\newcommand{\minigptfour}{{MiniGPT4}}
\newcommand{\fishdip}{{FISHDIP}}
\newcommand{\videochatrone}{{VideoChat-R1}}
\newcommand{\dlc}{{DLC}}
\newcommand{\cambrianone}{{Cambrian-1}}
\newcommand{\pearl}{{PEARL}}
\newcommand{\framefusion}{{FrameFusion}}
\newcommand{\rose}{{ROSE}}
\newcommand{\icd}{{ICD}}
\newcommand{\gem}{{GEM}}
\newcommand{\unilm}{{UNILM}}
\newcommand{\vcd}{{VCD}}
\newcommand{\llavagrounding}{{LLaVA-Grounding}}
\newcommand{\visionzip}{{VisionZip}}
\newcommand{\pmodmllm}{{p-MoD}}
\newcommand{\cswin}{{CSWin}}
\newcommand{\lavin}{{LaVIN}}
\newcommand{\pvp}{{PVP}}
\newcommand{\victor}{{Victor}}
\newcommand{\llavauhd}{{LLaVA-UHD}}
\newcommand{\corda}{{CorDA}}
\newcommand{\woodpecker}{{Woodpecker}}
\newcommand{\bioclip}{{BioCLIP}}
\newcommand{\marine}{{MARINE}}
\newcommand{\pali}{{PaLI}}
\newcommand{\sfllava}{{SF-LLaVA}}
\newcommand{\ritual}{{RITUAL}}
\newcommand{\setar}{{SeTAR}}
\newcommand{\ctts}{{CT$^2$S}}
\newcommand{\dor}{{DOR}}
\newcommand{\avisc}{{AvisC}}
\newcommand{\alcd}{{ALCD}}
\newcommand{\mustdrop}{{MustDrop}}
\newcommand{\logicrl}{{LogicRL}}
\newcommand{\pai}{{PAI}}
\newcommand{\lrv}{{LRV}}
\newcommand{\hio}{{HIO}}
\newcommand{\gtp}{{GTP}}
\newcommand{\moslora}{{MoSLoRA}}
\newcommand{\patch}{{PATCH}}
\newcommand{\vdgd}{{VDGD}}
\newcommand{\miniplm}{{MiniPLM}}
\newcommand{\coca}{{CoCA}}
\newcommand{\llavaalign}{{LLaVA-Align}}
\newcommand{\hydralora}{{HydraLoRA}}
\newcommand{\tcprompt}{{TCP}}
\newcommand{\atkd}{{ATKD}}
\newcommand{\minillm}{{MiniLLM}}
\newcommand{\fdivergence}{\textit{f}-DISTILL}
\newcommand{\emrmerging}{{EMR-Merging}}
\newcommand{\hiddenkey}{{HiddenKey}}
\newcommand{\smilellm}{{SMILE}}
\newcommand{\teamlora}{{TeamLoRA}}
\newcommand{\taskarithmetic}{{Task-Arithmetic}}
\newcommand{\sparsegpt}{{SparseGPT}}
\newcommand{\twinmerge}{{Twin-Merging}}
\newcommand{\tac}{{TAC}}
\newcommand{\besa}{{BESA}}
\newcommand{\dam}{{DAM}}
\newcommand{\dcc}{{DCC}}
\newcommand{\stepgrpo}{{StepGRPO}}
\newcommand{\ipt}{{IPT}}
\newcommand{\sparsevlm}{{SpareseVLM}}
\newcommand{\ties}{{TIES}}
\newcommand{\vila}{{VILA}}
\newcommand{\mft}{{MFT}}
\newcommand{\dobisvd}{{Dobi-SVD}}
\newcommand{\imccd}{{IMCCD}}
\newcommand{\lorahub}{{LoraHub}}
\newcommand{\vltrojan}{{VL-Trojan}}
\newcommand{\llavaphi}{{LLaVA-$\phi$}}
\newcommand{\vilatwo}{{VILA$^2$}}
\newcommand{\llamavid}{{LLaMA-VID}}
\newcommand{\videorone}{{Video-R1}}
\newcommand{\bliva}{{BLIVA}}
\newcommand{\sdft}{{SDFT}}
\newcommand{\multijail}{{MultiJail}}
\newcommand{\argue}{{ArGue}}
\newcommand{\llava}{{LLaVA}}
\newcommand{\dora}{{DoRA}}
\newcommand{\mofo}{{MoFO}}
\newcommand{\lisa}{{LISA}}
\newcommand{\llavahr}{{LLaVA-HR}}
\newcommand{\smop}{{SMoP}}
\newcommand{\loraplus}{{LoRA+}}
\newcommand{\sphinx}{{SPHINX}}
\newcommand{\honeybee}{Honeybee}
\newcommand{\vtw}{{VTW}}
\newcommand{\mpvit}{{MPViT}}
\newcommand{\cropa}{{CroPA}}
\newcommand{\spu}{{SPU}}
\newcommand{\tga}{{TGA}}
\newcommand{\lmbff}{{LM-BFF}}
\newcommand{\hallucidoctor}{{HalluciDoctor}}
\newcommand{\llama}{{LLaMA}}
\newcommand{\vicuna}{{Vicuna}}
\newcommand{\cpr}{{CPR}}
\newcommand{\spt}{{SPT}}
\newcommand{\coprompt}{{CoPrompt}}
\newcommand{\dare}{{DARE}}
\newcommand{\lth}{{LTH}}
\newcommand{\tasl}{{TaSL}}
\newcommand{\calm}{{CALM}}
\newcommand{\fish}{{FISH}}
\newcommand{\hacl}{{HACL}}
\newcommand{\wanda}{{Wanda}}
\newcommand{\maskllm}{{MaskLLM}}
\newcommand{\mthreeid}{{M3ID}}
\newcommand{\rlcf}{{RLCF}}
\newcommand{\stprompt}{{STP}}
\newcommand{\neglabel}{{NegLabel}}
\newcommand{\restore}{{RESTORE}}
\newcommand{\prompttuning}{{Prompt Tuning}}
\newcommand{\dpl}{{DPL}}
\newcommand{\ptp}{{PTP}}
\newcommand{\eazy}{{EAZY}}
\newcommand{\swarm}{{SWARM}}
\newcommand{\damp}{{DAMP}}
\newcommand{\fgrpr}{{FGRPR}}
\newcommand{\dpr}{{DPR}}
\newcommand{\saco}{{SaCo}}
\newcommand{\cpt}{{CPT}}
\newcommand{\qinsight}{{Q-Insight}}
\newcommand{\lori}{{LoRI}}
\newcommand{\cmpa}{{CMPA}}
\newcommand{\vars}{{VARS}}
\newcommand{\krona}{{Krona}}
\newcommand{\coda}{{CoDA}}
\newcommand{\lorapro}{{LoRA-Pro}}
\newcommand{\clipvip}{{CLIP-ViP}}
\newcommand{\vipllava}{{ViP-LLaVA}}
\newcommand{\ciat}{{CIAT}}
\newcommand{\sift}{{SIFT}}
\newcommand{\dits}{{DiTs}}
\newcommand{\aprompt}{{Aprompt}}
\newcommand{\absvit}{{AbSViT}}
\newcommand{\vlp}{{Vision-Language Pre-trained Models}}
\newcommand{\vlpab}{{VLPM}}
\newcommand{\san}{{SAN}}
\newcommand{\supermerge}{{SUPERMERGE}}
\newcommand{\clapaudio}{{CLAP}}
\newcommand{\rlt}{{RLT}}
\newcommand{\cascadeclip}{{Cascade-CLIP}}
\newcommand{\declip}{{DeCLIP}}
\newcommand{\imagebind}{{ImageBind}}
\newcommand{\uavod}{{UAV-OD}}
\newcommand{\clip}{{CLIP}}
\newcommand{\bop}{{BoP}}
\newcommand{\bicro}{{BiCro}}
\newcommand{\defo}{{DeFO}}
\newcommand{\melora}{{MeLoRA}}
\newcommand{\reslora}{{ResLoRA}}
\newcommand{\sharelora}{{ShareLoRA}}
\newcommand{\vera}{{VERA}}
\newcommand{\llavamerge}{{LLaVA-PruMerge}}
\newcommand{\flamingo}{{Flamingo}}
\newcommand{\ssam}{{SSAM}}
\newcommand{\alip}{{ALIP}}
\newcommand{\gps}{{GPS}}
\newcommand{\prefixtuning}{{PrefixTuning}}
\newcommand{\npt}{{NPT}}
\newcommand{\dac}{{DAC}}
\newcommand{\fsam}{{FSAM}}
\newcommand{\livt}{{LiVT}}
\newcommand{\scmoe}{{SCMoE}}
\newcommand{\instructblip}{{InstructBLIP}}
\newcommand{\blip}{{BLIP}}
\newcommand{\bliptwo}{{BLIP-2}}
\newcommand{\qwenvl}{{QWen-VL}}
\newcommand{\stwo}{{S$^{2}$}}
\newcommand{\alignVL}{{ALIGN}}
\newcommand{\ramt}{{R-AMT}}
\newcommand{\adamvmoe}{{AdaMVMoE}}
\newcommand{\tpt}{{TPT}}
\newcommand{\graphadapter}{{GraphAdapter}}
\newcommand{\lingualsmoe}{{Lingual-SMoE}}
\newcommand{\tai}{{TaI}}
\newcommand{\dapt}{{DAPT}}
\newcommand{\dualprompt}{{DualPrompt}}
\newcommand{\dept}{{DePT}}
\newcommand{\losparse}{{LoSparse}}
\newcommand{\fuller}{{FULLER}}
\newcommand{\depthclip}{{DepthCLIP}}
\newcommand{\vlab}{{VLAB}}
\newcommand{\lpt}{{LPT}}
\newcommand{\tvp}{{TVP}}
\newcommand{\promptkd}{{PromptKD}}
\newcommand{\metatrans}{{Meta-Transformer}}
\newcommand{\gasam}{{GA-SAM}}
\newcommand{\sprompt}{{SP}}
\newcommand{\promptstyler}{{PromptStyler}}
\newcommand{\umt}{{UMT}}
\newcommand{\pego}{{PEGO}}
\newcommand{\plot}{{PLOT}}
\newcommand{\doprompt}{{DoPrompt}}
\newcommand{\misa}{{MISA}}
\newcommand{\protext}{{ProText}}
\newcommand{\distllm}{{DISTLLM}}
\newcommand{\emotionllama}{{Emotion-LLaMA:}}
\newcommand{\vipt}{{ViPT}}
\newcommand{\kgcoop}{{KgCoOp}}
\newcommand{\bridge}{{Birdge}}
\newcommand{\dferc}{{DFERC}}
\newcommand{\xprompt}{{Xprompt}}
\newcommand{\relearn}{{ReLearn}}
\newcommand{\tailor}{{Tailor}}
\newcommand{\dualcoopplus}{{DualCoOP++}}
\newcommand{\upt}{{UPT}}
\newcommand{\itwomcl}{{I$^2$MCL}}
\newcommand{\gblending}{{G-Blending}}
\newcommand{\greedy}{{Greedy}}
\newcommand{\mmanet}{{MMANet}}
\newcommand{\grda}{{GRDA}}
\newcommand{\lmf}{{LMF}}
\newcommand{\kapt}{{KAPT}}
\newcommand{\mfm}{{MFM}}
\newcommand{\etwovpt}{{E$^2$VPT}}
\newcommand{\audioclip}{{AudioCLIP}}
\newcommand{\hotprotein}{{HotProtein}}
\newcommand{\prac}{{PRAC}}
\newcommand{\promptsrc}{{PromptSRC}}
\newcommand{\diffpurning}{{DiffPurning}}
\newcommand{\prograd}{{ProGrad}}
\newcommand{\pcb}{{PCB}}
\newcommand{\bike}{{BIKE}}
\newcommand{\maple}{{MaPLe}}
\newcommand{\fdmer}{{FDMER}}
\newcommand{\mtwopt}{{M2PT}}
\newcommand{\film}{{FiLM}}
\newcommand{\laclip}{{LaCLIP}}
\newcommand{\segzero}{{SegZero}}
\newcommand{\selfmm}{{Self-MM}}
\newcommand{\conc}{{Concatenation}}
\newcommand{\summ}{{Summation}}
\newcommand{\clipreid}{{CLIP-REID}}
\newcommand{\gated}{{Gated}}
\newcommand{\capforvideo}{{Cap4Video}}
\newcommand{\agm}{{AGM}}
\newcommand{\icode}{{i-Code}}
\newcommand{\msrl}{{MSRL}}
\newcommand{\mslr}{{MSLR}}
\newcommand{\fagm}{{FAGM}}
\newcommand{\cat}{{CAT}}
\newcommand{\omoe}{{OMoE}}
\newcommand{\taslplus}{{TaSLPLUS}}
\newcommand{\dualcoop}{{DualCoOp}}
\newcommand{\saf}{{SAF}}
\newcommand{\shape}{{SHAPE}}
\newcommand{\rigorllm}{{RigorLLM}}
\newcommand{\aptm}{{APTM}}
\newcommand{\fishr}{{Fishr}}
\newcommand{\soho}{{SOHO}}
\newcommand{\ijepa}{{I-JEPA}}
\newcommand{\pmr}{{PMR}}
\newcommand{\smil}{{SMIL}}
\newcommand{\man}{{MAN}}
\newcommand{\mbt}{{MBT}}
\newcommand{\dmd}{{DMD}}
\newcommand{\pmf}{{PMF}}
\newcommand{\tipadater}{{Tip-Adapter}}
\newcommand{\clipadapter}{{CLIP-Adapter}}
\newcommand{\coop}{{CoOP}}
\newcommand{\softmask}{{SoftMask}}
\newcommand{\clippo}{{CLIPPO}}
\newcommand{\cocoop}{{CoCoOp}}
\newcommand{\cafo}{{CaFo}}
\newcommand{\pcme}{{PCME}}
\newcommand{\ape}{{APE}}
\newcommand{\ogmge}{{OGMGE}}
\newcommand{\disf}{{DiSF}}
\newcommand{\vificlip}{{ViFi-CLIP}}
\newcommand{\fdt}{{FDT}}
\newcommand{\hpt}{{HPT}}
\newcommand{\shaspec}{{ShaSpec}}
\newcommand{\mtlora}{{MTLoRA}}
\newcommand{\cleanclip}{{CleanCLIP}}
\newcommand{\fourierft}{{FourierFT}}
\newcommand{\vilt}{{ViLT}}
\newcommand{\metaadapter}{{MetaAdapter}}
\newcommand{\gmc}{{GMC}}
\newcommand{\denseclip}{{DenseCLIP}}
\newcommand{\chils}{{CHiLS}}
\newcommand{\mtwofnet}{{M2FNet}}
\newcommand{\mmcosine}{{MMCosine}}
\newcommand{\lora}{{LoRA}}
\newcommand{\offsitetuning}{{Offsite-Tuning}}
\newcommand{\adapter}{{Adapters}}
\newcommand{\pst}{{PST}}
\newcommand{\lst}{{LST}}
\newcommand{\ltsft}{{LT-SFT}}
\newcommand{\moefication}{{MoEfication}}
\newcommand{\toast}{{TOAST}}
\newcommand{\fastv}{{FastV}}
\newcommand{\adalora}{{AdaLoRA}}
\newcommand{\pafi}{{PaFi}}
\newcommand{\less}{{LESS}}
\newcommand{\vpt}{{VPT}}
\newcommand{\infoprompt}{{InfoPrompt}}

\newcommand{\llm}{{Large Language Model}}
\newcommand{\llmabbrv}{{LLM}}
\newcommand{\mllm}{{Multimodal Large Language Model}}
\newcommand{\mllmabbrv}{{MLLM}}
\newcommand{\fm}{{Foundation Model}}
\newcommand{\fmabbrv}{{FM}}
\newcommand{\reinlearn}{{Reinforcement Learning}}
\newcommand{\reinlearnabbrv}{{RL}}

\newcommand{\grpo}{{Group Relative Policy Optimization}}
\newcommand{\grpoabbrv}{{GRPO}}
\newcommand{\ppo}{{Proximal Policy Optimization}}
\newcommand{\ppoabbrv}{{PPO}}

\newcommand{\ours}{{Mixed Advantage  Policy Optimization}}
\newcommand{\oursabbrv}{{MAPO}}

\newcommand{\firstmodule}{{Advantage Percent Deviation}}
\newcommand{\firstmdouleabbrv}{{APD}}
\newcommand{\secondmodule}{{Trajectory Certainty Reweight}}
\newcommand{\secondmoduleabbrv}{{TCR}}

\newcommand{\qwentwofivevlsevenins}{{Qwen2.5-VL-7B-Instruct}}

\newcommand{\geothreek}{{Geo3K}}
\newcommand{\emoset}{{EmoSet}}

\newcommand{\mathvista}{{MathVista}}
\newcommand{\wemath}{{WeMath}}
\newcommand{\mathverse}{{MathVerse}}
\newcommand{\webemo}{{WEBEmo}}
\newcommand{\emotionsix}{{Emotion6}}
\newcommand{\mathvision}{{MathVision}}

\newcommand{\vanilla}{{Vanilla}}

\newcommand{\tworow}[2]{
\begin{tabular}{@{}c@{}}
{#1}  \\
{#2}  \\
\end{tabular}
}

\section{Introduction}

Recent advances in the reasoning capabilities of \fm{} (\fmabbrv) {\cite{OpenAIOone_arXiv24,Kimik1_arXiv25, DeepSeekR1_arXiv25, LightROne_arXiv25, DeepSeekR1_arXiv25,Internvl35_arXiv25}} have been largely driven by improvements in long Chain of Thought (CoT) generation. Among various enhancement strategies, \reinlearn{} (\reinlearnabbrv{}) {\cite{HumanFeedback_NeurIPS22, GPT4_arXiv23, Deepseekmath_arXiv24, OpenReasoner_arXiv25, ReinforceRej_arXiv25}} has emerged as a powerful post-training technique, enabling \fmabbrv{} to refine their CoT reasoning through self-improvement. Therefore, \reinlearnabbrv{} serves as the key mechanism for unlocking the reasoning ability in various domains.

Notably, \grpo{} (\grpoabbrv{}) {\cite{Deepseekmath_arXiv24}} is introduced as a popular reinforcement strategy. \grpoabbrv{} generates and refines a group of reasoning paths through the group-relative advantage estimation based on rule-based reward functions.  Thus, a key difference with traditional reinforcement methods, such as proximal policy optimization {\cite{PPO_arXiv17}} and direct preference optimization {\cite{DPO_NeurIPS23,DICE_arXiv24,DPOSurvey_arXiv25}}, is that \grpoabbrv{} eliminates the need for an additional learned reward critic model, instead leveraging efficient sampling from the \fm{} policy. Witnessing the success of \grpo{}, its advantage function plays a key role in promoting trajectories with relatively higher advantages, thereby guiding the policy model to update towards more reliable directions. Despite recent advancements, \grpoabbrv{} and its variants generally maintain a fixed advantage formulation throughout the entire training cycle {\cite{DeepSeekR1_arXiv25,ShareGRPO_ICCV25,DeepSeekR1_arXiv25}}. However, this approach overlooks a significant challenge: \textit{the fixed advantage fails to provide meaningful signals for samples with varying trajectory certainty degrees}. 

\begin{figure*}[t]
\centering
\begin{center}
\includegraphics[width=\linewidth]{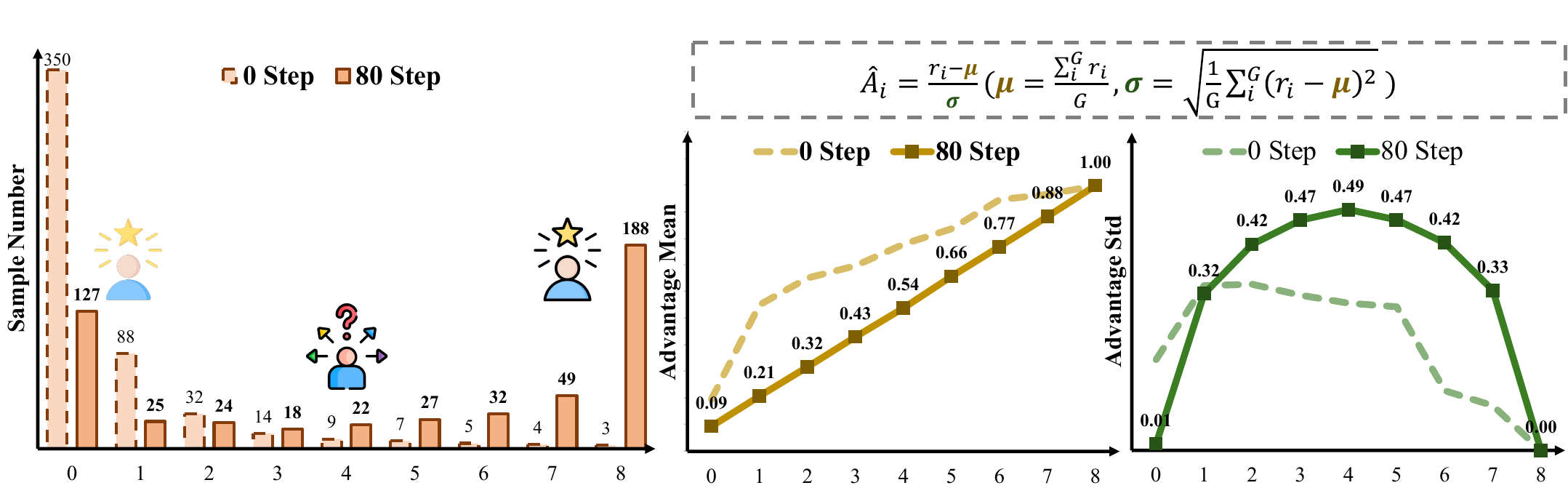}
\end{center}
\vspace{-10pt}
\captionsetup{font=small}
\caption{\textbf{Observation}. During the reinforcement, different samples appear with diverse successful trajectory numbers $N \! =\! \sum_{i=1}^{G} \mathbf{1}_{\{r_i = 1\}}$ (\textbf{X-axis}). Samples with lowest trajectory certainty (\protect\includegraphics[scale=0.03,valign=c]{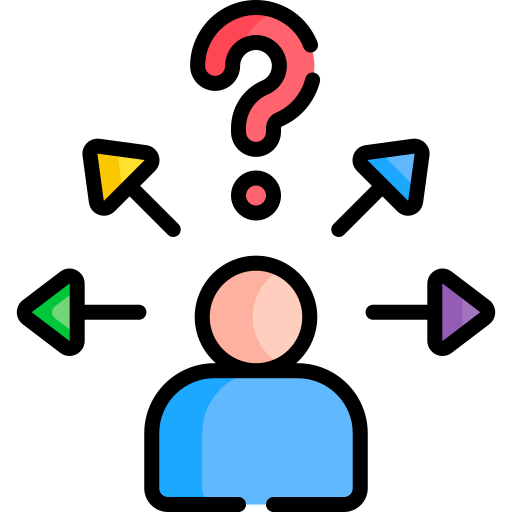}) tends to achieve most diverse prediction pattern, \ie, $N =4$. Experiments are conducted on the \geothreek{} with rollout number $G=8$.} 
\label{fig:motivation}
\vspace{-20pt}
\end{figure*}

To analyze the drawbacks of existing advantage formulations, we first define \textbf{\textit{trajectory certainty}} within the sampling group. The advantage is computed from verifiable rewards, typically format and accuracy metrics, to jointly measure the trajectory score {\cite{DeepSeekR1_arXiv25,Deepseekmath_arXiv24,ShareGRPO_ICCV25,StepGRPO_arXiv25,NoisyRollout_arXiv25,SPO_arXiv25}}. For a sampled trajectory, we declare the \textbf{success} only if it achieves the correct answer on all reward metrics. Consequently, each trajectory can be viewed as a Bernoulli trial with outcome: failure or success. Then, in the group sampling, the number of successes over repeated draws follows a binomial distribution, and \textit{high-certainty samples tend to yield nearly identical outcomes across draws}, \ie, samples that are too hard or too easy. We then formally derive the definition of trajectory certainty as follows:
\begin{tcolorbox}[
enhanced,                
sidebyside,              
colframe=black!70,       
colback=yellow!5,        
boxrule=1pt,             
arc=4mm,                 
lefthand width=0.06\linewidth,  
sidebyside gap=5mm,      
top=0.2mm,               
bottom=0.2mm,            
boxsep=0.5mm,            
]
\hspace*{-0.2cm}
\includegraphics[width=1.\linewidth]{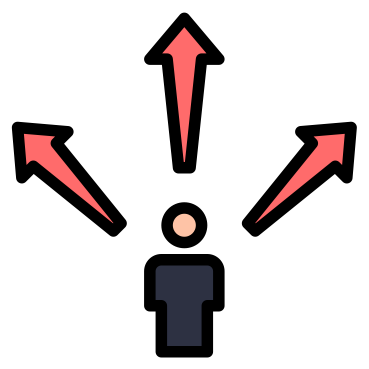}
\tcblower 
\textit{\textbf{Trajectory Certainty in \grpoabbrv{}}: High certainty  \protect\includegraphics[scale=0.04,valign=c]{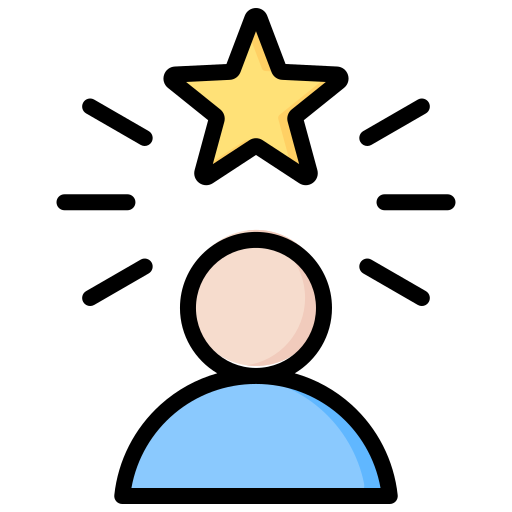} corresponds to trajectories with lower prediction variance, while low certainty \protect\includegraphics[scale=0.04,valign=c]{Figure/confusion.png} reflects higher variance.}
\end{tcolorbox}

We analyze the sample behavior in \cref{fig:motivation} and reveal two underlying limitations of the existing advantage paradigm. \textit{First}, \textbf{\textit{Advantage Reversion}}: high-certainty samples may receive more differentiated advantage allocations than low-certainty ones. Specifically, a high-certainty sample (\protect\includegraphics[scale=0.03,valign=c]{Figure/self-confidence-1.png}) with rewards $\bm{r}_{\text{High}} \!=\! \{0.9, 1.0, 1.0, 1.0\}$ receives a more discriminative advantage allocation than a low-certainty sample (\protect\includegraphics[scale=0.03,valign=c]{Figure/confusion.png}) with $\bm{r}_{\text{Low}} \!=\! \{0.1, 0.9, 1.0, 1.0\}$ due to {\color{DarkGreen}\textbf{a small advantage standard deviation $\sigma$}}. However, high-certainty samples do not require strong penalization, whereas low-certainty trajectories benefit from stronger correction. \textit{Second}, \textbf{\textit{Advantage Mirror}}: high-certainty samples (\protect\includegraphics[scale=0.03,valign=c]{Figure/self-confidence-1.png} \& \protect\includegraphics[scale=0.03,valign=c]{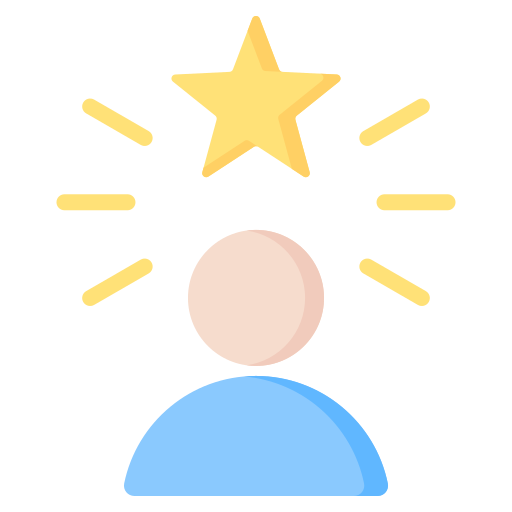}) also require distinct advantage allocation for extreme cases. In particular, the existing advantage formulation does not take into account the {\color{DarkYellow}\textbf{monotonic advantage scores $\mu$}} and therefore treats easy and hard samples indistinguishably. The core issue is that the same advantage formulation cannot be applied uniformly across samples with different trajectory certainty. In summary, this motivates us to rethink the advantage design and decomposes it into two sub-questions: \hypertarget{Q1}{\textbf{\expandafter{\romannumeral1})}} \textit{how to design the advantage function for high-certainty samples}? and \hypertarget{Q2}{\textbf{\expandafter{\romannumeral2})}} \textit{how to adaptively combine advantage functions for samples with varying trajectory certainty}?

To address the question \myhyperlink{Q1}{\textbf{\expandafter{\romannumeral1})}}, we introduce the \firstmodule{} (\firstmdouleabbrv{}), which replaces the advantage from standard z-score normalization to relative normalization. Specifically, the original advantage formulation is expressed as $\hat{A}_i = \frac{r_i - {\color{DarkYellow}\mu}}{{\color{DarkGreen}\sigma}}$. For high-certainty sample trajectories, this formulation fails to capture the overall level of reward scores. Besides, variance in the rollout trajectory can yield a small $\sigma=\text{std}(\bm{r})$, which in turn leads to numerical instability and uncontrollable boundary on advantage allocation. To deal with this drawback, we introduce a novel advantage function for high-certainty samples as $\hat{A}_i^{\firstmdouleabbrv{}} \! = \! \frac{r_i -{\color{DarkYellow}\mu}}{{\color{DarkYellow}\mu}}$. Regarding question \myhyperlink{Q2}{\textbf{\expandafter{\romannumeral2})}}, we propose the \secondmodule{} (\secondmoduleabbrv{}) to determine the sample advantage function based on trajectory certainty. Inspired by Bernoulli sampling, each trajectory is treated as either a success or a failure. A trajectory group exhibits the highest uncertainty when the success-to-failure ratio is fifty percent. Therefore, we use trajectory certainty to dynamically reweight the advantage function from $\hat{A}_i$ to $\hat{A}_i^{\firstmdouleabbrv{}}$. In this work, we argue that the existing advantage formulation is not consistently appropriate for samples with varying levels of trajectory certainty. To address this issue, we propose a simple yet effective method, \textbf{M}ixed \textbf{A}dvantage \textbf{P}olicy \textbf{O}ptimization (\textbf{MAPO}), which rethinks the advantage formulation in \grpoabbrv{}. To validate our approach, we conduct extensive experiments across multiple datasets using the Qwen2.5-VL-7B architecture, demonstrating the superior performance in both In-Domain and Out-of-Domain aspects. Our contributions are summarized as follows:

\begin{fullitemize}

\item We focus on the \grpo{} paradigm and reveal that  existing advantage formulation faces two unavoidable challenges: advantage reversion and advantage mirror.

\item We propose \ours{} (\oursabbrv{}), a simple yet effective method to overcome existing advantage limitations. Preliminary, we introduce trajectory certainty to evaluate sample behavior. We propose \firstmodule{} for high-certainty advantage estimation and utilize \secondmodule{} to dynamically construct the advantage function.

\item We perform a comprehensive analysis on reasoning scenarios, including mathematics and emotion fields. Through a series of ablation studies, the promising results empirically validate the effectiveness of the proposed mixture advantage strategies in enhancing \grpoabbrv{} overall performance.

\end{fullitemize}

\section{Related Works}

\subsection{Foundation Model}
The development of \llm{} (\llmabbrv{}) has revolutionized artificial intelligence, significantly transforming the way machines understand and generate human language. Notable examples of \llmabbrv{} include the GPT series {\cite{GPT2_OPENAI19, GPT3_NeurIPS20, GPT4_arXiv23}}, Meta LLaMA {\cite{LLaMA_arXiv23}}, and Google PaLM {\cite{PaLM_arXiv22, PaLMv2_arXiv23}}, all of which have demonstrated impressive capabilities in natural language understanding and generation. These advancements have sparked considerable interest in extending \llmabbrv{} to handle multi-modal inputs, particularly by incorporating vision components, which has led to the development of \mllm{} (\mllmabbrv{}). Building on the success of \llmabbrv{}, growing interest has emerged in constructing end-to-end \mllm{} (\mllmabbrv{}) systems, such as \flamingo{} {\cite{Flamingo_NeurIPS22}}, \bliptwo{} {\cite{BLIP_ICML22, BLIPv2_ICML23}}, \instructblip{} {\cite{InstructBLIP_NeurIPS23}}, \qwenvl{} {\cite{QwenVL_arXiv23}}, \llava{} {\cite{LLaVA_NeurIPS23, LLaVA15_CVPR24, LLaVAPhi_arXiv24,LLaVANext_arXiv24}}, and \vila{} {\cite{VILA_CVPR24,VILA2_arXiv24,NVILA_CVPR25}}. Existing \mllmabbrv{} solutions typically rely on visual extractors {\cite{CLIP_ICML21,ViT_ICLR21,dino_ICCV21}} to encode visual features, using a connector module to project visual tokens into the word embedding space of the \llmabbrv{}, \ie, treating visual input as a foreign language {\cite{BEiT3_CVPR23}}. Subsequently, the visual and textual tokens are concatenated and fed into the \llmabbrv{}. The \llmabbrv{} is then used to perform various vision-language tasks in an auto-regressive manner. As a result, foundation models are gradually evolving from a single-textual modality to multimodal capabilities. However, existing works predominantly focus on supervised fine-tuning (SFT) on large-scale pre-training datasets. The success of OpenAI o1 {\cite{GPT2_OPENAI19,OpenAIOone_arXiv24,HumanFeedback_NeurIPS22}} highlights the powerful potential of reinforcement learning in post-training to enhance model reasoning capabilities. With the open-sourcing of Deepseek-R1 {\cite{DeepSeekR1_arXiv25}} and Qwen {\cite{Qwen_arXiv23,QwenVL_arXiv23,Qwen2.5_arXiv25,Qwen3_arXiv25}}, reasoning models are now widely deployed locally, drawing attention from the research community to the efficiency of long chain-of-thought generation for foundation models. And utilizing the reinforcement technique to empower foundation models with reasoning capabilities has emerged as a pivotal methodology beyond the limitations of SFT.

\subsection{\grpo{}}
\label{sec:relatedworksgrpo}
 Several methods have been proposed to elicit reasoning abilities on mathematical and scientific problems, enabling foundation models to better handle inference and analysis. Especially, \grpo{} (\grpoabbrv{}) has recently garnered significant attention in the research field, as its rule-based reward function effectively enhances the reasoning capabilities of large models. Existing exploration or variants of \grpoabbrv{} could normally be divided into the following streams. \ding{182} \textit{Think Trajectory Diversity}.  This paradigm focuses on diversifying the thinking process to facilitate a more meaningful candidate rollout. Specifically, it boosts the trajectory from two angles: input perturbation and process polish. First, constructing the data augmentation technique for \mllm{} to enhance both the quantity and quality of training data. \noisyrollout{} \cite{NoisyRollout_arXiv25} leverages the noise annealing schedule to construct the noisy image text pairs. \visonpert{} {\cite{VisionMatters_arXiv25}} introduces three targeted perturbations: distractor concatenation, dominance-preserving mixup, and random rotation. \sharegrpo{} {\cite{ShareGRPO_ICCV25}} turns to expand the question space for a given question via data transformation.
Second, polishing the thinking process acts as a reliable direction to monitor the thinking behavior. \stepgrpo{} {\cite{StepGRPO_arXiv25}} requires the think process to explicitly reveal key intermediate steps.  Both \sophicvlrone{}  \cite{SophiaVLR1_arXiv25} and \grpocare{} \cite{GRPOCARE_arXiv25} utilize an external thinking reward model that evaluates the quality of the entire thinking process. \mgrpo{} {\cite{MGRPO_arXiv25}} recycles previous think messages for self-correction learning. \hintgrpo{} {\cite{HintGRPO_ICCV25}} adaptively provides hints to the samples.  However, the aforementioned solutions require constructing dedicated data augmentation strategies or modifying the thinking process, which introduces additional computational costs or an external thinking reward evaluation model. \ding{183}  \textit{Reward Formulation Refinement}. With respect to verifiable reward construction {\cite{Deepseek_arXiv24,Kimik1_arXiv25}}, it is the predefined rules and normally incorporates the Format Reward and Accuracy Reward. The former requires the model output should meet the required HTML tag format of  \texttt{<think>} and  \texttt{<answer>}. The latter is determined by comparing the model output class with the ground truth class, yielding a value of 1 for correct classification and 0 for incorrect classification. Thus, recent works design different verifiable reward functions for different specific tasks. For instance, \vrft{} \cite{VRFT_arXiv25} proposes the intersection over union reward for object detection. \visionreasoner{} {\cite{VisionReasoner_arXiv25}} introduces diverse perception rewards in a unified framework. Both \grpolambda{} \cite{GRPOLambda_arXiv25} and \grpolead{} {\cite{GRPOLEAD_arXiv25}} consider the over-length penalty reward. However, this pattern typically focuses on adapting to specific tasks, which limits cross-task generalization. Additionally, it requires careful tuning of hyperparameter reward weights for different reward metrics. Therefore, this pattern fails to achieve robust performance across diverse real-world application settings. \ding{184}  \textit{Advantage Estimation Redesign}. Towards advantage estimation, recent researches investigate better trajectory importance measurement via reformulation or rescaling operation. \drgrpo{} \cite{DrGRPO_arXiv25} and \gpg{}  \cite{GPG_arXiv25} consider removing the standard deviation to alleviate the reward bias. \seedgrpo{} \cite{SEEDGRPO_arXiv25} reweights the advantages based on the semantic entropy to measure the output uncertainty.  \krpo{} {\cite{KRPO_arXiv25}} introduces a lightweight Kalman filter approach for accurate advantage estimation. But this paradigm faces the hyperparameter selection dilemma, or consistent advantages for different samples. We conclude the weakness of existing \grpoabbrv{} variants in \cref{tab:grpo_weakness}. In our work, we reveal that \textit{samples appear distinct trajectory certainty behavior and utilizing  uniform advantage strategy unavoidably degrades partial samples optimization}. Therefore, we dynamically set the advantage function based on the trajectory certainty to boost the overall reinforcement effect.

\begin{table*}[t]\small
\captionsetup{font=small}
\caption{\textbf{Weakness} for different \grpoabbrv{} variants. Refer to \cref{sec:relatedworksgrpo} for details. }
\label{tab:grpo_weakness}
\vspace{-10pt}
\centering
\resizebox{\textwidth}{!}{
\setlength\tabcolsep{3pt}
\renewcommand\arraystretch{1.2}
\begin{tabular}{r||c|c|c|c}
\hline\thickhline
\rowcolor{gray!20}
\textbf{Methods}  &  \textbf{Input Space Augmentation} & \textbf{Think Cost Increase} & \textbf{Specific Task Adaption} & \textbf{Additional Hyper-Parameter} \\
\hline\hline
\multicolumn{4}{l}{\textbf{\textit{Think Trajectory Diversity}}~\raisebox{-0.2em}{\includegraphics[height=1em]{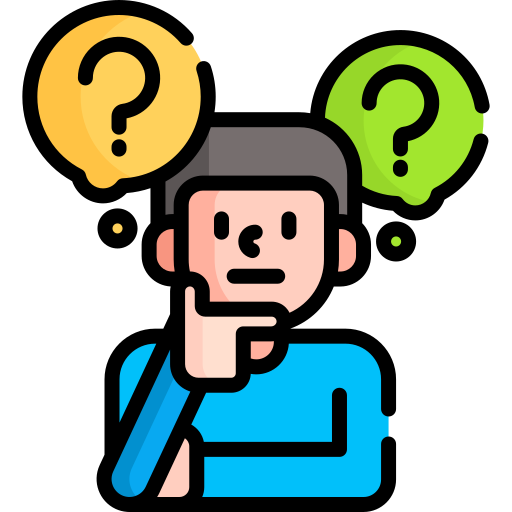}}} \\  
\hline
\rowcolor{rxklightblue}
\noisyrollout  & {\color{rxkdarkdarkblue}\Checkmark{}}  (Noisy Distortion) &  & & {\color{rxkdarkdarkblue}\Checkmark{}} (Initial Noise Strength) \\
\visonpert  & {\color{rxkdarkdarkblue}\Checkmark{}}  (Visual Augmentation)   &  & & {\color{rxkdarkdarkblue}\Checkmark{}}  (Perturbation Types)\\
\rowcolor{rxklightblue}
\sharegrpo  & {\color{rxkdarkdarkblue}\Checkmark{}}  (Textual Enrichment) &  & & {\color{rxkdarkdarkblue}\Checkmark{}}  (Question Variants Number)  \\
\stepgrpo  & & {\color{rxkdarkdarkblue}\Checkmark{}} (Step Think) & & {\color{rxkdarkdarkblue}\Checkmark{}} (Key Steps Number) \\
\rowcolor{rxklightblue}
\grpocare    & & {\color{rxkdarkdarkblue}\Checkmark{}} (Reference Model) & & {\color{rxkdarkdarkblue}\Checkmark{}}  (Consistency Coefficient) \\
\hline\hline
\multicolumn{4}{l}{\textbf{\textit{Reward Formulation Refinement}~\raisebox{-0.2em}{\includegraphics[height=1em]{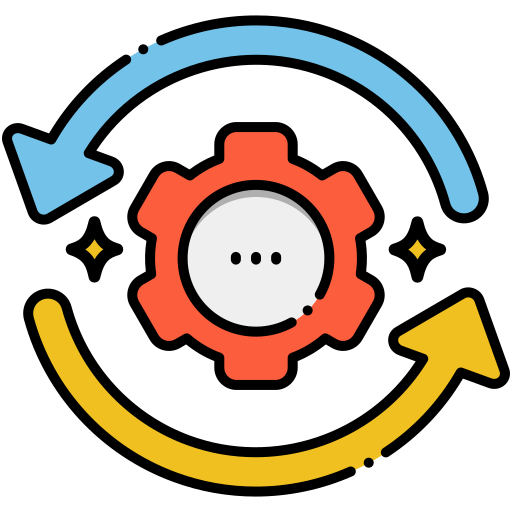}}}} \\  
\hline
\rowcolor{rxklightblue}
\vrft  & & & {\color{rxkdarkdarkblue}\Checkmark{}} (Visual IoU Reward) &\\
\grpolambda  & & & {\color{rxkdarkdarkblue}\Checkmark{}}  (Length Penalty)  & {\color{rxkdarkdarkblue}\Checkmark{}} (Top-$\lambda$ Fraction)\\
\rowcolor{rxklightblue}
\grpolead{} & & & {\color{rxkdarkdarkblue}\Checkmark{}}  (Length Reward)  & {\color{rxkdarkdarkblue}\Checkmark{}} (Advantage Rescale Factor)\\
\hline\hline
\multicolumn{4}{l}{\textbf{\textit{Advantage Estimation Redesign}}~\raisebox{-0.2em}{\includegraphics[height=1em]{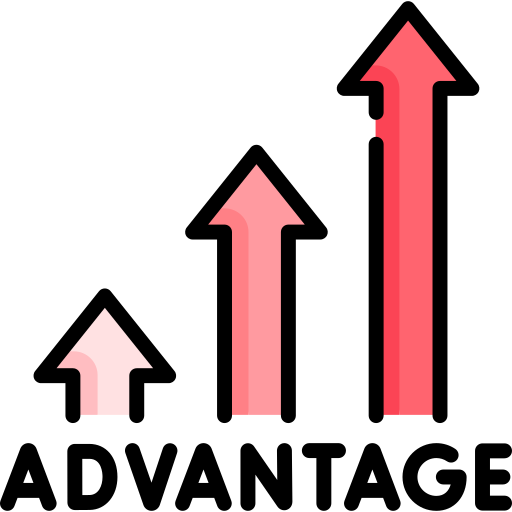}}} \\  
\hline
\rowcolor{rxklightblue}
\seedgrpo{} & & &  & {\color{rxkdarkdarkblue}\Checkmark{}} (Advantage Rescale Factor)\\
\gpg{} & & &  & {\color{rxkdarkdarkblue}\Checkmark{}} (Valid Sample Threshold)\\
\hline
\thickhline
\end{tabular}}
\vspace{-15pt}
\end{table*}

\section{Methodology}

\subsection{Preliminary}
\label{sec:preliminary}

\grpo{} (\grpoabbrv{}) {\cite{Deepseekmath_arXiv24}} is a variant of Proximal Policy Optimization (PPO) {\cite{PPO_arXiv17}} originally developed to enhance mathematical reasoning in \llmabbrv{}. However, it can also be effectively adapted to improve visual reasoning in \mllm{}. \grpoabbrv{} begins by constructing the current policy model $\pi_\theta$ and a reference model $\pi_{old}$, where the latter represents the ``old" policy or the policy from a previous iteration. Let $\rho_Q$ denote the distribution of prompts or questions. Given a prompt $q \sim \rho_Q$, \grpoabbrv{} samples a group of outputs ${o_1, o_2, \ldots, o_G}$ from the old model $\pi_{old}$. It then optimizes the policy model $\pi_\theta$ by maximizing the following objective function:
\begin{equation} \setlength\abovedisplayskip{0pt} \setlength\belowdisplayskip{0pt}
\label{eq:optimize}
\mathcal{J}_{\mathrm{GRPO}}(\theta) =  \mathbb{E}_{q \sim \rho_{\mathcal{Q}}}  \mathbb{E}_{o \sim \pi_{old} (\cdot |q)} \left[\frac{1}{G}\sum_i^G f_\epsilon \left( \frac{\pi_{\theta}(o_i|q)}{\pi_{old}(o_i|q)}, \hat{A}_i \right) \right]-\beta \mathbb{D}_{KL}[\pi_\theta || \pi_{ref}],
\end{equation}
where $\beta$ is the hyper-parameter. $f_\epsilon(x, y) = \min(xy, \text{clip}(x, 1-\epsilon, 1 + \epsilon)y)$. $\hat{A}_i$ is the advantage calculated based on the relative rewards of the outputs inside each group. To be precise, for each question $q$, a group of outputs $\{o_1, o_2, \ldots, o_G\}$ are sampled from the old
policy model $\pi_{old}$. A reward function ($R$) is then used to score the outputs, yielding $G$ rewards $\bm{r} = \{r_1, r_2,\ldots, r_G \}$ correspondingly, where $r_i = R(q,o_i)$.  The  mean reward is then calculated as $\mu = \frac{1}{G} \sum_{i=1}^G r_i$ and the standard deviation is defined as $\sigma = \sqrt{\frac{1}{G} \sum_{i=1}^{G} \left(r_i - \mu \right)^2 }$. The default normalized advantage for the
$i^{th}$ rollout is defined as the following formulation:
\begin{equation}
\setlength\abovedisplayskip{0pt} \setlength\belowdisplayskip{0pt}
\label{eq:originaladvantage}
\hat{A}_i=\frac{r_i-\mu}{\sigma}.
\end{equation}

\subsection{Proposed Method}
\label{sec:proposedmethod}
\noindent \textbf{Observation}.  We model the trajectory outcome as a Bernoulli random variable, \(X \sim \text{Bernoulli}(p), \; X \in \{0,1\}\), where \(X=1\) denotes a successful trajectory and \(X=0\) denotes a failure. The success probability \(p\) is defined by the expectation of \(X\), \(\mathbb{E}[X] = p\), and the variance of this distribution is \(\text{Var}(X) = p(1-p)\), which quantifies the certainty of the trajectory outcome, shown in \cref{fig:motivation}. However, directly measuring \(p\) is challenging, so we estimate it empirically using the ratio \(p \approx \frac{N}{G}\), where \(G\) is the total number of sampled trajectories. \(N\) is the number of successful trajectories and is defined as the following formulation:
\begin{equation} \setlength\abovedisplayskip{0pt} \setlength\belowdisplayskip{0pt}
\label{successful_trajectory}
N = \sum_{i=1}^{G} \mathbf{1}_{\{r_i = 1\}}.
\end{equation}
This empirical estimation approximates the true probability \(p\) via observed trajectories. Thus, we reveal that \textit{samples exhibit varying certainty level within the \grpoabbrv{} sampling process}.

\begin{figure*}[t!]
\centering
\begin{center}
\includegraphics[width=\textwidth]{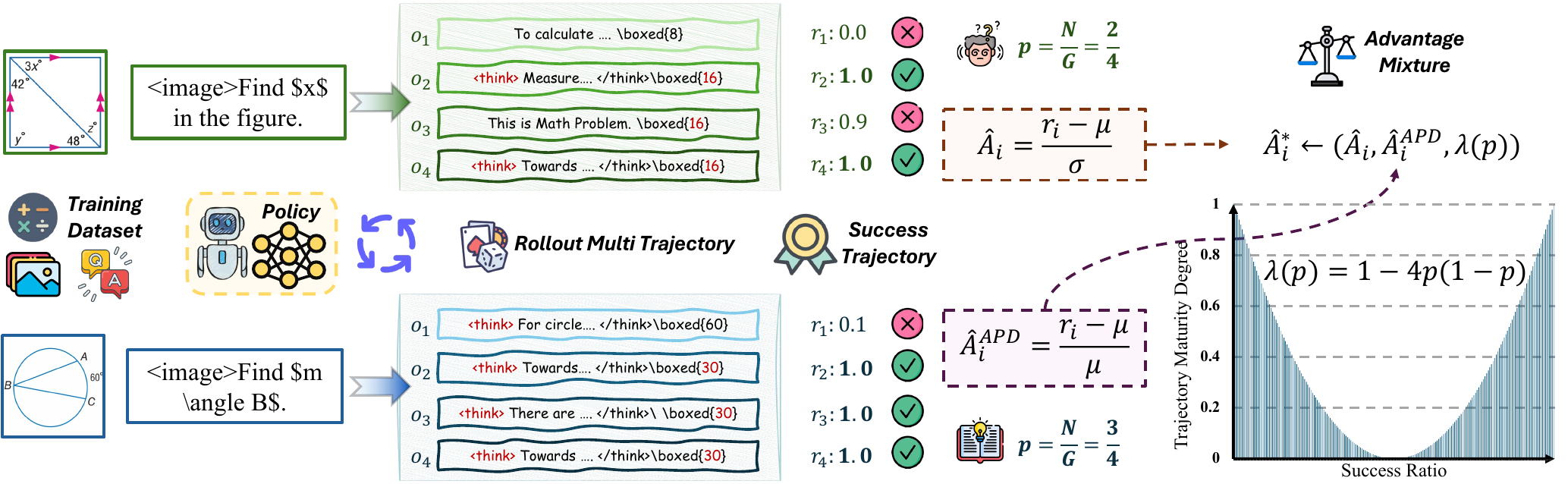}
\put(-152,70){\scriptsize{\cref{eq:originaladvantage}}}
\put(-72,92){\scriptsize{\cref{eq:mixadvantage}}}
\put(-152,46){\scriptsize{\cref{eq:proposedadvantage}}}
\end{center}
\vspace{-10pt}
\captionsetup{font=small}
\caption{\textbf{Architecture illustration of \oursabbrv{}}. We reveal that the trajectory certainty varies across samples. In general, we introduce the \firstmodule{} to replace the advantage function for high-certainty elements. We utilize the \secondmodule{} to dynamically reweight the advantage function via trajectory certainty. Assume rollout number $G\!=\!4$. Best viewed in color. Zoom in for details. See \cref{sec:proposedmethod}.}
\label{fig:framework}
\vspace{-15pt}
\end{figure*}

\noindent \textbf{\firstmodule}. Sample would appear high certainty, when the prediction variance is close to zero ($\text{Var}(X) \rightarrow 0 $), \ie, $p \rightarrow 0$ or $p \rightarrow 1$, which typically corresponds to overly easy or difficult instances. In such cases, existing advantage faces two key challenges: \textit{Advantage Reversion} and \textit{Advantage Mirror}. Specifically, the advantage formulation, $\hat{A}_i = \frac{r_i - \mu}{\sigma}$, can produce misleading behaviors between trajectories with high and low certainty. For instance, in the case of \textit{Advantage Reversion}, the high-certain trajectory with a relatively high reward of $0.9$ in a batch of $\bm{r}_{\text{High}}\!=\![0.9,\,1,\,1,\,1]$ is assigned a large negative advantage ($\min \hat{A}_i \!=\! -1.73$), which is more extreme than the low-certain ones like $\bm{r}_{\text{Low}}\!=\![0.1,\,0.1,\,1,\,1]$, due to the small standard deviation exaggerating deviations from the mean. Similarly, as for \textit{Advantage Mirror}, two reward batches that are symmetric around the center, such as $[0,\,0.1,\,0.1,\,0.1]$ and $[0.9,\,1,\,1,\,1]$, yield mirrored normalized advantage scores $[-1.73,\,0.57,\,0.57,\,0.57]$, which makes semantically distinct cases to appear structurally equivalent normalization. These examples show how reliance on $\mu$ and $\sigma$ alone can distort the relative evaluation of trajectories, especially when the variance is abnormally small or the rewards are symmetrically distributed, thus echoing a more robust advantage function. 

Therefore, in our work, to address the question \myhyperlink{Q1}{\textbf{\expandafter{\romannumeral1})}}: \textit{high-certainty samples advantage reconstruction}, we introduce the {\firstmodule{} (\firstmdouleabbrv{})}, to effectively address the issues of \textit{Advantage Reversion} and \textit{Advantage Mirror}. Instead of relying on z-score normalization, \firstmdouleabbrv{} measures the relative deviation of each trajectory reward from the batch mean reward, formulated as follows:
\begin{equation} \setlength\abovedisplayskip{0pt} \setlength\belowdisplayskip{0pt} 
\label{eq:proposedadvantage}
\hat{A}_i^{\firstmdouleabbrv}= \frac{r_i - \mu}{\mu}.
\end{equation}
This design emphasizes the proportional difference between individual rewards and the central tendency, ensuring that the advantage reflects not only the relative ordering but also the magnitude of deviation in percentage terms. By doing so, \firstmdouleabbrv{} mitigates the instability caused by abnormally small standard deviations and prevents mirrored advantage allocation from being treated as equivalent, thereby providing a more stable and reasonable trajectory quality evaluation.

\noindent \textbf{\secondmodule}. Considering that samples are in various trajectory certainty conditions, it is essential to dynamically adjust the advantage formulation across different samples. We propose the \secondmodule{} (\secondmoduleabbrv{}), which adaptively reconstructs the advantage function based on trajectory certainty to address \myhyperlink{Q2}{\textbf{\expandafter{\romannumeral2})}} \textit{various-certainty sample advantage reweighting}. This design ensures that sample-specific characteristics are preserved, leading to a more faithful and stable evaluation of trajectory quality. 

To be precise, we formalize \secondmoduleabbrv{} by introducing a certainty-aware weighting scheme. The key intuition is that when a trajectory exhibits high uncertainty (immature stage), the advantage should rely more on a variance-sensitive formulation (\cref{eq:originaladvantage}), while in highly certain (mature) stages it should instead emphasize a mean-relative formulation (\cref{eq:proposedadvantage}) that remains stable even when variance collapses. To operationalize this idea, we use the estimated trajectory certainty $p$ to interpolate these two advantages for different samples. We denote the trajectory certainty degree as follows:
\begin{equation} \setlength\abovedisplayskip{0pt} \setlength\belowdisplayskip{0pt}
\label{eq:trajeorycertainty}
\lambda(p)= 1 - 4p(1-p) \in [0,1] \quad (p \in [0,1]).
\end{equation}
And, then we further construct the sample-wise advantage construction as follows:
\begin{equation} \setlength\abovedisplayskip{0pt} \setlength\belowdisplayskip{0pt}
\label{eq:mixadvantage}
\hat{A}_i^* =  (1-\lambda(p)) * \underbrace{\tcbhighmath[colback=green!8]{\frac{r_i-\mu}{\sigma}}}_{{\color{DarkGreen}\textbf{Deviation-based}}} + \lambda(p) * \underbrace{\tcbhighmath[]{\frac{r_i-\mu}{\mu}}}_{{\color{DarkYellow}\textbf{Mean-based}}}.
\end{equation}
The standard deviation–based advantage is weighted by $1-\lambda(p)$, while the complementary factor $\lambda(p)$ is assigned to the mean-based advantage. In this way, the contribution shifts smoothly from deviation-based signals under uncertainty to mean-based signals under certainty, ensuring a balanced and robust construction of the advantage function across different trajectory certainty levels. Thus, we replace the original advantage $\hat{A}_i$ to proposed $\hat{A}_i^*$ in \cref{eq:mixadvantage} in \cref{eq:optimize} for optimization.
As a result, our method reveals the trajectory certainty phenomenon and effectively mitigates existing advantage limitations via dynamical advantage reweight operation. We provide the methodological framework in \cref{fig:framework} and the algorithm description in \cref{alg:ours}.

\begin{figure*}[t]
\noindent\begin{minipage}[]{0.67\textwidth}
\centering
\small{
\resizebox{\textwidth}{!}{
\setlength\tabcolsep{2pt}
\renewcommand\arraystretch{1.2}
\begin{tabular}{r||c|c|c}
\hline\thickhline
\rowcolor{gray!20}
&  & \multicolumn{2}{c}{{Advantage Problem}}    \\
\rowcolor{gray!20}
\multirow{-2}{*}{{Methods}}  & \multirow{-2}{*}{{Advantage Formulation}}  & {\color{DarkGreen}{Adv. Reversion}} & {\color{DarkYellow}{Adv. Mirror}}   \\
\hline\hline
\rowcolor{rxklightblue}
\grpoabbrv{} \hypertarget{A1}{\ding{168}}  &  $\hat{A}_i=\frac{r_i-\mu}{\sigma}$  &  \Checkmark{} & \Checkmark{}  \\
\drgrpo{} \hypertarget{A2}{\ding{169}}   & $\hat{A}_i=r_i-\mu$ & & \Checkmark{} \\
\rowcolor{rxklightblue} 
\gpg{} \hypertarget{A3}{\ding{170}}  & $\hat{A}_i= \alpha * (r_i-\mu)$ & & \Checkmark{} \\
\treerpo{} \hypertarget{A4}{\ding{171}}  &  $\hat{A}_i=\frac{r_i-\mu}{\mu (1-\mu)}$ & & \Checkmark{} \\
\hline
\rowcolor{rxklightblue} 
\oursabbrv{} \hypertarget{A5}{\ding{72}} & $\hat{A}_i= \lambda \frac{r_i-\mu}{\sigma} \!+\! (1-\lambda)  \frac{r_i-\mu}{\mu} $  & - & -\\
\hline
\thickhline
\end{tabular}}}
\end{minipage}
\hfill
\hspace{-5pt}
\noindent\begin{minipage}[]{0.32\textwidth}
\centering
\includegraphics[width=\linewidth]{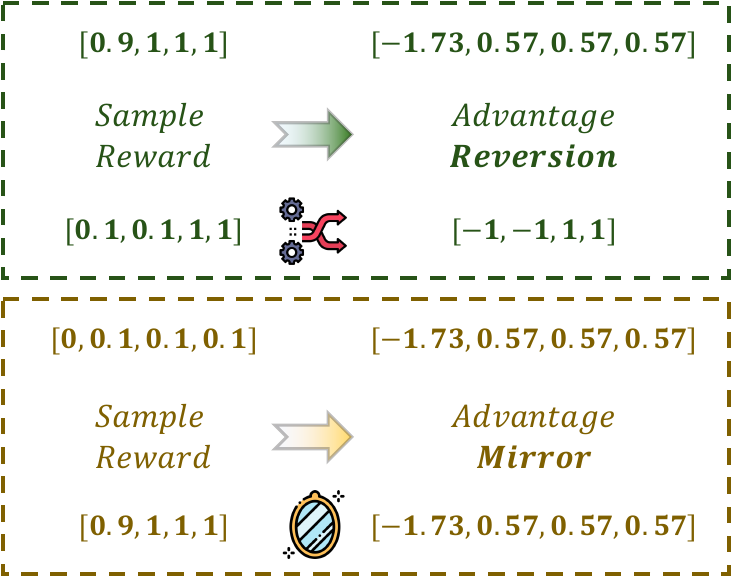}
\end{minipage}
\vspace{-10pt}
\captionsetup{font=small}
\caption{\textbf{Discussion on existed advantage functions}. $\bm{r}$ means the group reward. We denote $\mu \!=\! mean(\bm{r})$ and $\sigma \!=\! std(\bm{r})$. The reward is defined as a combination of the format reward ($r_{Format}$) and the accuracy reward ($r_{Accuracy}$), with a weighting factor of $\beta = 0.9$, \ie, $r = (1-\beta) r_{Format} + \beta{} r_{Accuracy}$. $\alpha\!=\!0.6$ is the valid sample rescale parameter for \gpg{} \cite{GPG_arXiv25}. Refer to \cref{sec:discussionandlimitation} for details.
}
\vspace{-15pt}
\label{fig:advantagediscussion}
\end{figure*}

\begin{figure*}[t]
\centering
\begin{center}
\includegraphics[width=0.9\textwidth]{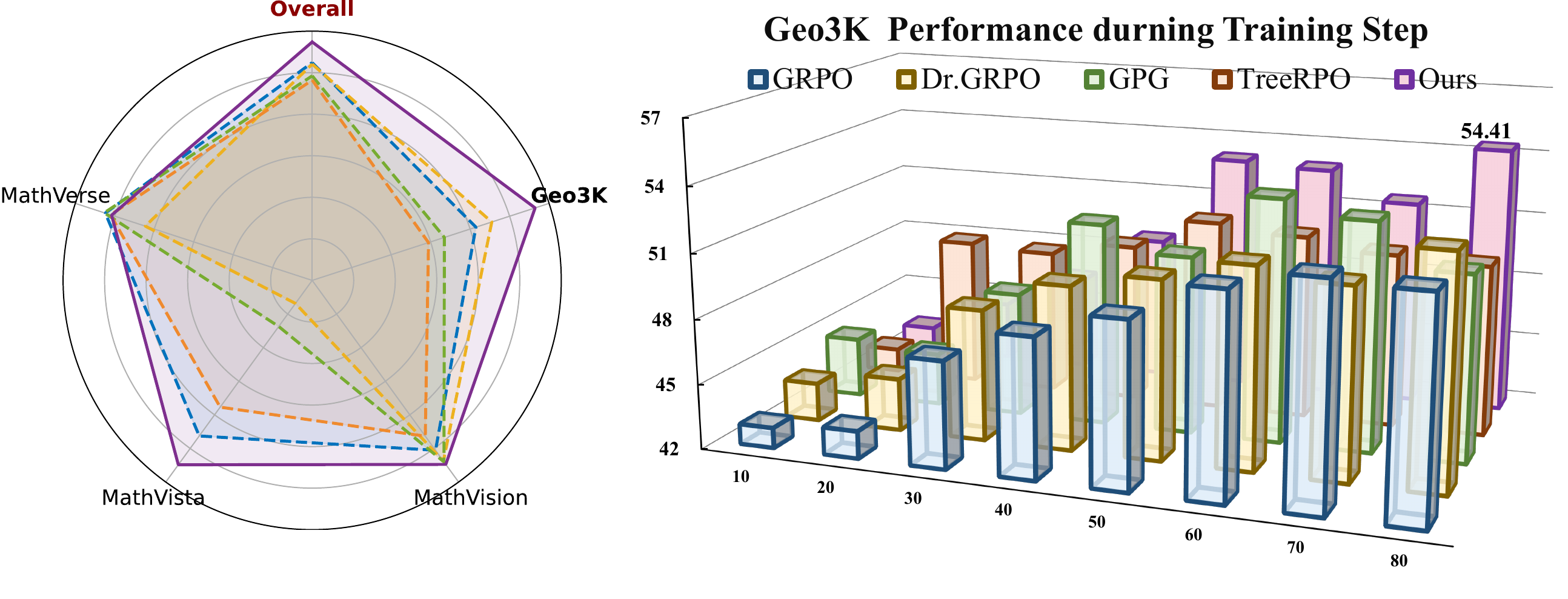}
\put(-182, 108){\footnotesize{\myhyperlink{A1}{\ding{168}}}}
\put(-142, 108){\footnotesize{\myhyperlink{A2}{\ding{169}}}}
\put(-102, 108){\footnotesize{\myhyperlink{A3}{\ding{170}}}}
\put(-67, 108){\footnotesize{\myhyperlink{A4}{\ding{171}}}}
\put(-37, 108){\footnotesize{\myhyperlink{A5}{\ding{72}}}}
\end{center}
\vspace{-20pt}
\captionsetup{font=small}
\caption{\textbf{Performance comparison with different advantage formulations} on the geometry task based on \qwentwofivevlsevenins{} model with rollout number $G\!=\!12$. Please refer to \cref{sec:discussionandlimitation} for details.}
\label{fig:advtangeformulations}
\vspace{-15pt}
\end{figure*}
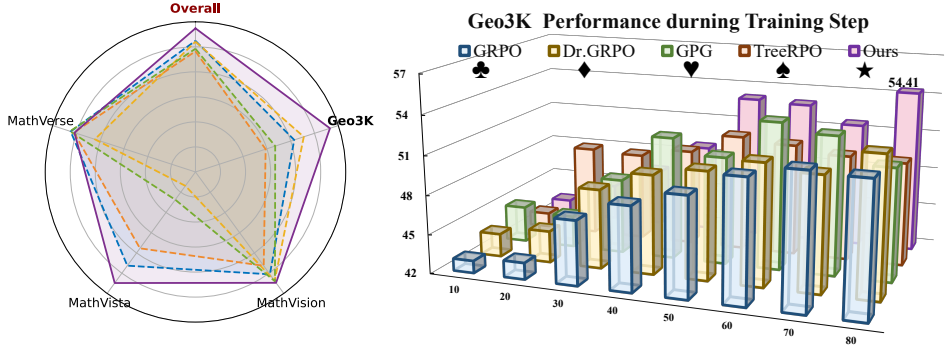

\subsection{Discussion and Limitation}
\label{sec:discussionandlimitation}

\noindent \textbf{Advantage Exploration}.  
The advantage function typically relies on group-based estimation from trajectory rewards $r_i \in \bm{r}$. Existing explorations {\cite{DrGRPO_arXiv25,GPG_arXiv25,TreeRPO_arXiv25}} can be broadly classified as the following directions. \textbf{\textit{First}}, methods such as \drgrpo{} {\cite{DrGRPO_arXiv25}}, \gpg{} {\cite{GPG_arXiv25}}, \sgrpo{} \cite{SGRPO_arXiv25}, and \wdone{} \cite{wd1_arXiv25} remove the standard variance normalization term to alleviate reward bias. More recently, \cite{TreeRPO_arXiv25} identifies that conventional normalization fails to scale advantages properly under continuous rewards, and proposes to rewrite the variance as $\sigma = \mu (1-\mu)$ to address this issue. Meanwhile, \krpo{} \cite{KRPO_arXiv25} introduces a lightweight Kalman filter to dynamically estimate latent reward mean and variance, enabling more adaptive advantage normalization.  However, these approaches fail to resolve both  \textit{advantage reversion} and \textit{advantage mirror} problems in \cref{fig:advantagediscussion}. Moreover, existing methods typically employ a uniform advantage formulation across all samples, overlooking the uniqueness of individual sample conditions. In contrast, we observe that samples exhibit varying degrees of certainty during optimization. Motivated by this insight, we propose a novel advantage function that leverages relative deviation for high-certainty samples, and further introduce a certainty-aware reweighting scheme that dynamically adjusts the advantage construction based on trajectory certainty. This design ensures a more faithful and stable evaluation of the sample situation across diverse training conditions. We further conduct the empirical experiments to validate the proposed mixture advantage solutions in \cref{fig:advtangeformulations} and \oursabbrv{} achieves a satisfying performance.

\noindent \textbf{Conceptual Difference}. Utilizing \grpoabbrv{} to enhance foundation models with reasoning capabilities has gained significant attention  \cite{DrGRPO_arXiv25,GRPOCARE_arXiv25,SFTMemorizeRLGeneralize_arXiv25,ReLIFT_arXiv25}. Recent studies have explored the contribution of sample selection to the effectiveness of \grpoabbrv{} across three main streams. First, one line focuses on highlighting relatively simple samples to achieve a stable optimization. For instance, \seedgrpo{} \cite{SEEDGRPO_arXiv25} utilizes semantic entropy to measure answer diversity and applies more conservative updates to hard questions. However, blindly emphasizing easy samples can restrict the model exploration ability \cite{DoesRLincentiveReasoning_arXiv25} and face model entropy collapse {\cite{PFPPO_arXiv24}}. Second, another paradigm seeks to highlight hard samples. For example, \grpolead{} amplifies learning signals for challenging problems using a difficulty-aware advantage reweight. Recently, \cite{HardExamplesGRPO_arXiv25} has pointed out that the hardest examples consistently yield superior performance on reasoning benchmarks.  However, this approach leads to longer convergence times. The third group focuses on eliminating meaningless samples. \dapo{} {\cite{DAPO_arXiv25}} and \gpg{} \cite{GPG_arXiv25} aim to discard samples with vanishing advantages, \ie, $\sigma = 0$. \dapo{} considers over-sampling and filtering out prompts with mean accuracy $\mu\in \{0,1\}$, but this operation is not efficient in terms of training time. This inefficiency arises because the time required to collect a batch of desired examples is uncontrollable and depends on the task difficulty. In contrast, \gpg{} seeks more accurate gradient estimation by rescaling the gradient based on the valid samples ratio, with a validity threshold of $0.6$. In summary, existing work conducts a \textbf{monotonic emphasis} based on sample difficulty, which inevitably faces the prisoner dilemma of sample difficulty. In contrast, our work considers trajectory certainty and allocates different mixture ratios for high- and low-certainty samples, thereby introducing a \textbf{discriminative emphasis}. We reveal the gradient of \oursabbrv{} compared with \grpoabbrv{}. \textit{Without loss of generality}, we simplify the gradient analysis by
ignoring clipping and KL regularization and considering the reward as a Bernoulli variable.
We define the ratio between the gradients of \oursabbrv{} and \grpoabbrv{} as:
\begin{equation}
\label{eq:ratio}
\varrho(p)\triangleq \frac{\nabla_\theta \mathcal J_{\mathrm{MAPO}}}{\nabla_\theta \mathcal J_{\mathrm{GRPO}}}
=(1-\lambda(p))+\lambda(p)\sqrt{\frac{1-p}{p}},\quad \lambda(p)=1-4p(1-p).
\end{equation}
By further analysis (see details in \cref{sec:theoryanalysis}), we obtain the following formulation:
\begin{equation}
\begin{cases}
\varrho(p) > 1, & p\in(0,\frac12),\\[4pt]
\varrho(p) = 1, & p = \frac{1}{2},\\[4pt]
0 < \varrho(p) < 1, & p\in(\frac12,1).
\end{cases}
\end{equation}
This shows that the mixed reward, \oursabbrv{} 
\textit{implicitly assigns larger gradients to harder samples (with $p<\frac12$) and smaller gradients to easier ones (with $p>\frac12$)}, which aligns with prior insights that appropriately emphasizing difficult samples enhances the performance \grpoabbrv{} {\cite{HardExamplesGRPO_arXiv25}}.

\noindent \textbf{Limitation}. Despite achieving satisfactory performance with free hyperparameters, our research has several limitations. First, our approach uses trajectory certainty to treat different samples selectively. In extreme reinforcement scenarios or when foundational model capabilities are limited, it becomes difficult to generate a diverse set of successful trajectories, as rollout may consistently fail. In such cases, our method could reduce to a single function strategy. Second, although our method  assigns different reward mechanisms for samples with different trajectory maturity levels, a more refined reward allocation method is still worth exploring. Third, due to computational constraints, our experiments are limited to models with up to 7B parameters and datasets with  a few thousand samples. Future work would aim to extend these findings to larger-scale scenarios.

\section{Experiments}

\subsection{Experimental Setup}
\label{sec:Setup}

\noindent \textbf{Environment and Datasets}.  We conduct experiments on two reasoning scenarios: mathematics and emotion. For training, we utilize \geothreek{} \pub{arXiv'21} {\cite{GeoThreeK_arXiv21}} and \emoset{} \pub{ICCV'23} {\cite{EmoSet_ICCV23}}. These two datasets are respectively comprised of $2.1K$ training samples. Furthermore, for the out-of-domain validation, we respectively adopt out-of-domain datasets in the math and emotion fields:
\mathvista{} \pub{arXiv'23} {\cite{MathVista_arXiv23}},  \mathvision{} \pub{NeurIPS'24} {\cite{MathVision_NeurIPS24}}, \mathverse{} \pub{ECCV'24} {\cite{MathVerse_ECCV24}}, \webemo{} \pub{ECCV'18} {\cite{WEBEmo_ECCV18}}, and \emotionsix{} \pub{CVPR'15} \cite{EmotionSix_CVPR15}. We provide a detailed dataset illustration in \cref{sec:datasetintrouction}.

\noindent \textbf{Architecture and Counterparts}. We utilize the popular open-source \qwentwofivevlsevenins{} as the base (Vanilla) model, which exhibits strong foundational capabilities well-suited for
subsequent \reinlearnabbrv{} training {\cite{Qwen2.5_arXiv25,QwenVL2.5_arXiv25}}. We further conduct the comparison with the \grpoabbrv{} {\cite{Deepseekmath_arXiv24}} and \dapo{} \cite{DAPO_arXiv25} to validate the effectiveness of our method.

\noindent \textbf{Implementation Details}. Experiments are conducted on 8 A100 GPUs. Detail is in \cref{sec:trainingdetails}.

\noindent \textbf{Evaluation Metrics}.
We evaluate both in-domain ($\mathcal{A}^\mathcal{S}$) and out-of-domain ($\mathcal{A}^\mathcal{T}$). Let $\mathcal{T}=\{\mathcal{T}_t\}_{t=1}^{|\mathcal{T}|}$ represent the unseen dataset set and $\mathcal{S}$ denote the training distribution. Thus, we derive the following evaluation metrics forms $ \mathcal{A}^\mathcal{S} = \text{Acc.}(\mathcal{S})$ and $\mathcal{A}^\mathcal{T} = \frac{1}{|\mathcal{T}|} \sum_t^{|\mathcal{T}|} \text{Acc.}(\mathcal{T}_i)$. Acc. denotes the accuracy metric. Furthermore, we use the Average metric to evaluate overall performance as $\bar{\mathcal{A}}= \frac{\mathcal{A}^\mathcal{S} +  \mathcal{A}^\mathcal{T}}{2}$.

\begin{table*}[t]\small
\captionsetup{font=small}
\caption{\textbf{Ablative study of key modules} for \oursabbrv{} via\qwentwofivevlsevenins{} with rollout number $G\!=\!12$.  Incorporate \textbf{sole} \firstmodule{} (\firstmdouleabbrv{}) can be regarded as the advantage replacement. Involving \textbf{both} \firstmdouleabbrv{} and \secondmoduleabbrv{} achieves a satisfying performance. Please refer to \cref{sec:ablation}. for details.}
\label{tab:ablation_loss}
\vspace{-10pt}
\centering
{
\resizebox{\columnwidth}{!}{
\setlength\tabcolsep{3pt}
\renewcommand\arraystretch{1.2}
\begin{tabular}{cc||c|cccIcc||c|ccIcc}
\hline \thickhline
\rowcolor{gray!20}
\firstmdouleabbrv{} & \secondmoduleabbrv  & \geothreek{}  & \mathvision{}  & \mathvista{}& \mathverse{} &  $\mathcal{A}^\mathcal{T}$ & $\bar{\mathcal{A}}$
& \emoset{} & \webemo{} & \emotionsix{} &  $\mathcal{A}^\mathcal{T}$ & $\bar{\mathcal{A}}$
\\
\hline\hline
\multicolumn{2}{c||}{Vanilla} 
& - & - & - & - & - & - & - & - & - & - & - 
\\ 
\hdashline
& &  51.91 &26.74 &	72.70 &	43.93 	& 47.79 & 49.85 
& 75.50 	&49.90 &	60.44 &	55.17&	65.33
\\ 
\ding{51} & 
& 50.92 & 	26.61 &	73.00 &	44.59 & 48.07 & 49.49 
& 77.20 &	50.55 &	61.95 &	56.25 & 66.72
\\ 
\ding{51} & Rand
& 53.41 & 24.90 & 71.20 & 43.38 & 46.49 & 49.95 & 76.90 & 50.20 & 62.12 & 56.16 & 66.53 

\\
\rowcolor{rxklightblue}
\ding{51} & \ding{51}
& \textbf{54.41} & 27.30 & 73.20 & 43.81 & \textbf{48.10} & \textbf{51.26} 
& \textbf{77.86} & 50.75 & 60.61 & \textbf{55.68}	& \textbf{66.77}\\
\hline
\thickhline
\end{tabular}}}
\vspace{-10pt}
\end{table*}

\begin{table*}[t]
\captionsetup{font=small}
\caption{\textbf{Performance comparison with \grpoabbrv{} variants} on the geometry and emotional reasoning tasks. We mark the {Best} in bold across different methods.  Refer to \cref{sec:compSOTA}.}
\label{tab:compare_sota_grpo_variants}
\vspace{-10pt}
\centering
\scriptsize{
\resizebox{\linewidth}{!}{
\setlength\tabcolsep{2.pt}
\renewcommand\arraystretch{1.2}
\begin{tabular}{r||c|cccIcc||c|ccIcc}
\hline\thickhline
\rowcolor{gray!20}
{Methods}  
& \geothreek{}  & \mathvision{}  & \mathvista{}& \mathverse{} &  $\mathcal{A}^\mathcal{T}$ & $\bar{\mathcal{A}}$
& \emoset{} & \webemo{} & \emotionsix{}  &  $\mathcal{A}^\mathcal{T}$ & $\bar{\mathcal{A}}$
\\
\hline\hline
\vanilla{} 
& 37.43 & 24.51 & 67.20  & 40.02  & 43.91 & 40.67
& 53.65 & 46.85 & 52.19 & 49.52 & 51.58
\\
\hdashline
\multicolumn{12}{l}{\textcolor{gray!100}{\textit{{\qwentwofivevlsevenins{} with Rollout Number $G=12$}}}} 
\\ 
\grpoabbrv{} 
& 51.91 &	26.74 &	72.70 &	43.93 	& 47.79 & 49.85 
& 77.20  &	49.90 &	60.44 &	55.17	& 66.18

\\

\dapo{} 
& 52.91 &	26.51 	& 73.50 &	44.59 & 48.20  &50.56 
& 76.05 &	50.60 	& 60.61 &	55.60 &	65.82
\\

\hdashline
\rowcolor{rxklightblue}
\oursabbrv{} 
& \textbf{54.41} &	27.30 &	73.20 &	43.81 &	48.10  & \textbf{51.26} 
& \textbf{77.86} &	50.75 &	60.61 &	55.68 &	\textbf{66.77} \\
\hline\hline
\multicolumn{12}{l}{\textcolor{gray!100}{\textit{{\qwentwofivevlsevenins{} with Rollout Number $G=8$}}}} 
\\ 
\grpoabbrv{} 
& 50.92 &	26.38 &	72.60 &	43.45 &	\textbf{47.48} & 49.20 
& 76.40 &	49.90 &	60.27 &	55.08 &	65.74
\\

\dapo{} 
& 50.42 &	26.41 &	72.40 &	43.15 &	47.32 & 48.87 
& 68.44 & 47.80 & 58.08 & 52.94 & 60.69
\\

\hdashline
\rowcolor{rxklightblue}
\oursabbrv{} 
& \textbf{54.24} &	27.37 &	71.30 &	43.40 &	47.36 &	\textbf{50.80}
& \textbf{77.46} & 50.05 & 61.28 & \textbf{55.66} & \textbf{66.56} \\ 
\hline
\thickhline
\end{tabular}}}
\vspace{-10pt}
\end{table*}

\begin{figure*}[t]
\centering
\subfigure[\small{\geothreek{} with $G=12$}]
{\includegraphics[width=0.50\textwidth]{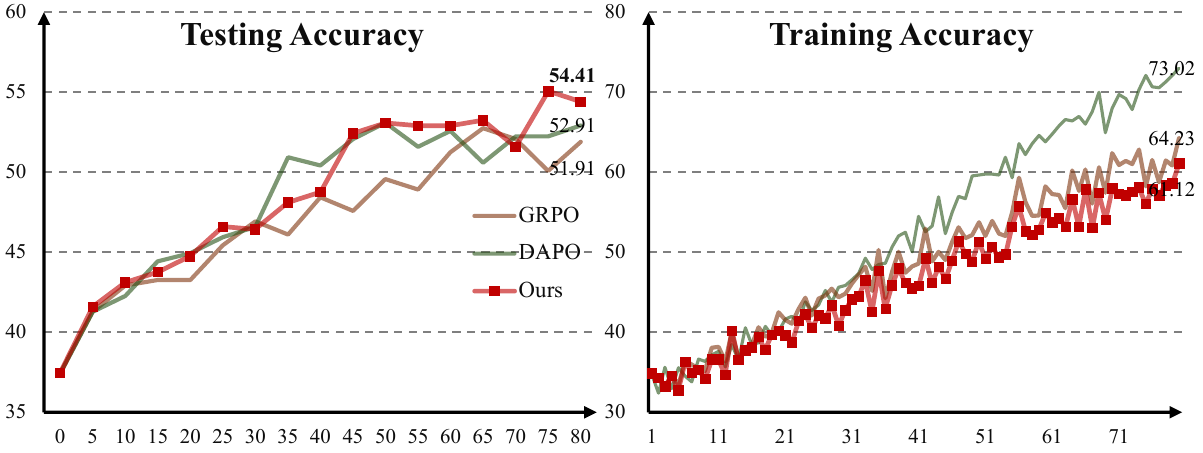}}
\hspace{-5pt}
\subfigure[\small{\geothreek{} with $G=8$}]
{\includegraphics[width=0.50\textwidth]{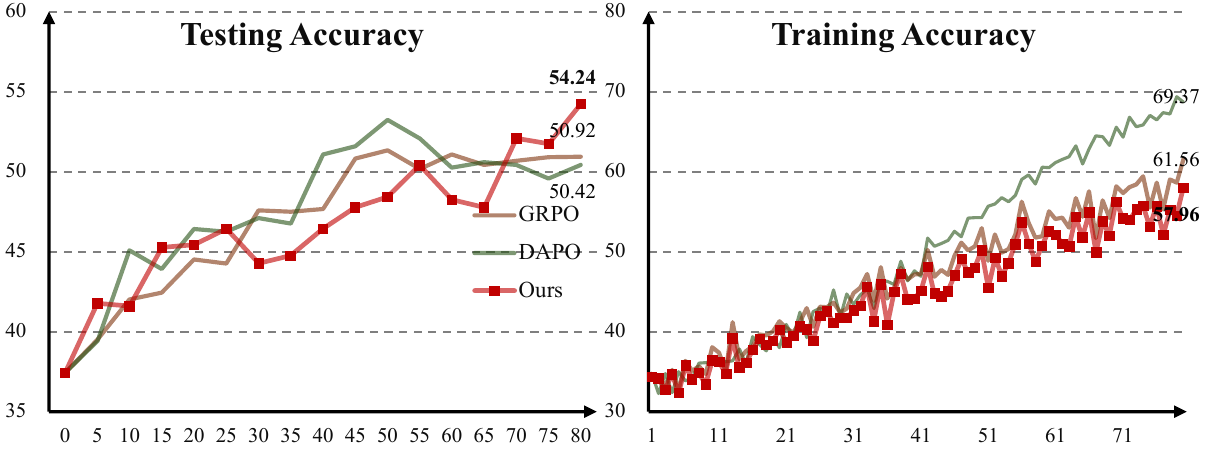}}
\vspace{-15pt}
\subfigure[\small{\emoset{} with $G=12$}]
{\includegraphics[width=0.50\textwidth]{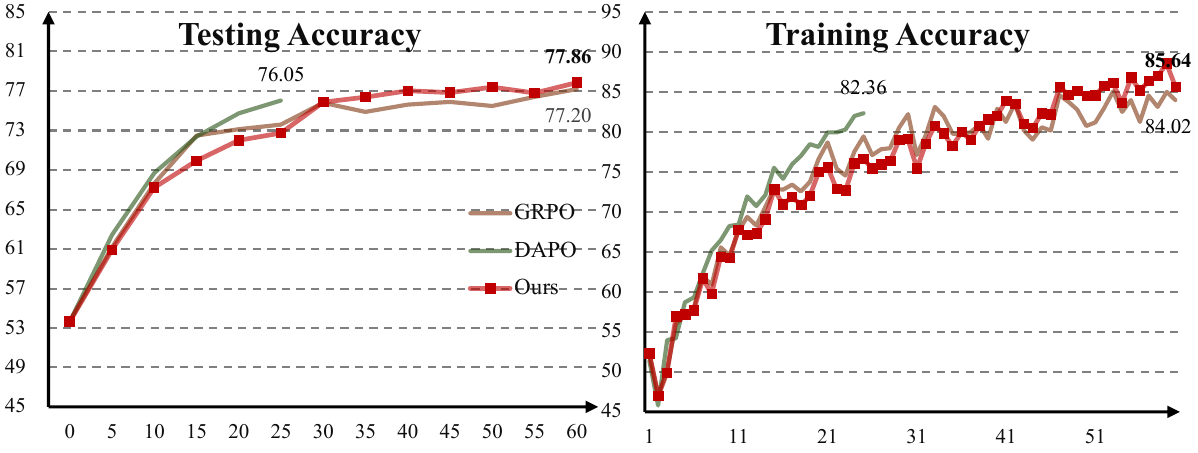}}
\hspace{-5pt}
\subfigure[\small{\emoset{} with $G=8$}]
{\includegraphics[width=0.50\textwidth]{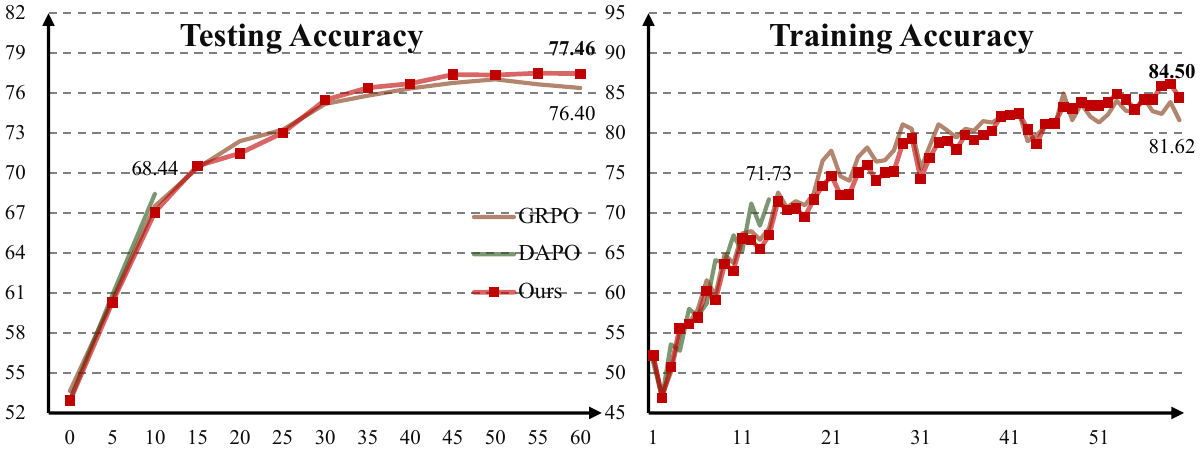}}
\captionsetup{font=small}   
\caption{\textbf{Training and testing accuracy during the training process}. \dapo{} fails to conduct complete training due to dynamic sampling failure in \emoset{} scenario. Refer to \cref{sec:compSOTA}.}
\label{fig:sota_epoch}
\vspace{-10pt}
\end{figure*}

\subsection{Diagnostic Analysis}
\label{sec:ablation}

We perform ablation studies on the \geothreek{} and \emoset{} datasets, utilizing the \qwentwofivevlsevenins{} model to facilitate an in-depth analysis. We quantitatively analyze the proposed \ours{} (\oursabbrv{}) in \cref{tab:ablation_loss}. The ablation demonstrates that solely replacing the advantage function from the original  $\hat{A}_i\!=\!\frac{r_i-\mu}{\sigma}$ to $\hat{A}_i^{\firstmdouleabbrv{}} \! = \! \frac{r_i -\mu}{\mu}$ leads to limited performance improvement or even degradation, which underscores the necessity for a dynamic advantage function. Furthermore, utilizing random weight allocation ($\lambda(p)\sim \mathcal{U}(0, 1)$ in \cref{eq:mixadvantage}) fails to achieve stable performance improvements. Thus, incorporating the \secondmodule{}, which accounts for trajectory certainty, further enhances overall performance.

\subsection{Comparison to State-of-the-Arts}
\label{sec:compSOTA}

We benchmark \oursabbrv{} against state-of-the-art reinforcement learning frameworks in reasoning tasks. As illustrated in \cref{tab:compare_sota_grpo_variants}, \oursabbrv{} consistently outperforms \vanilla{}, \grpoabbrv{}, and \dapo{} across both in-domain, \eg, \geothreek{} and out-of-domain, \eg, \mathvision{}, \mathvista{}, and \mathverse{}, showing strong generalization performance under different rollout numbers. With $G=12$, it achieves the highest overall accuracies ($51.26$ on math and $66.77$ on emotion), and even with $G=8$, it maintains superior results (50.80 and 66.55), validating that its mixed advantage formulation effectively mitigates advantage reversion and mirror issues while ensuring reliable optimization. Overall, it advances state-of-the-art performance with consistent gains across diverse reasoning tasks.

\section{Conclusion}
In our work, we focus on \grpo{} (\grpoabbrv{}) and observe that the advantage function plays a crucial role in evaluating trajectory importance. However, existing advantage formulations face two challenges: advantage reversion and advantage mirror. To address these issues, we propose \ours{} (\oursabbrv{}). In particular, we uncover the trajectory certainty property and introduce advantage percent deviation for high-certainty trajectories. Furthermore, we dynamically reweight the advantage function according to trajectory certainty, thereby adaptively tailoring the advantage to sample-specific characteristics. Our method offers three key advantages: First, {\textit{No Architecture Dependency}}: \oursabbrv{} operates without additional model architectures, ensuring high transferability across different architectures. Second, {\textit{No Thinking Pattern Conflict}}: our approach directly evaluates trajectory advantages while maintaining compatibility with diverse reasoning formats. Third, {\textit{No Hyper-Parameter Configuration}}: by leveraging trajectory certainty to adaptively reweight sample advantage formulations, our method avoids the need for additional hyperparameters, thereby improving reinforcement effectiveness. \oursabbrv{} has been validated across diverse scenarios, underscoring its potential for broader applications.



\bibliography{iclr2026_conference}

\begin{thebibliography}{76}
\providecommand{\natexlab}[1]{#1}
\providecommand{\url}[1]{\texttt{#1}}
\expandafter\ifx\csname urlstyle\endcsname\relax
  \providecommand{\doi}[1]{doi: #1}\else
  \providecommand{\doi}{doi: \begingroup \urlstyle{rm}\Url}\fi

\bibitem[Alayrac et~al.(2022)Alayrac, Donahue, Luc, Miech, Barr, Hasson, Lenc, Mensch, Millican, Reynolds, Ring, Rutherford, Cabi, Han, Gong, Samangooei, Monteiro, Menick, Borgeaud, Brock, Nematzadeh, Sharifzadeh, Bi\'{n}kowski, Barreira, Vinyals, Zisserman, and Simonyan]{Flamingo_NeurIPS22}
Jean-Baptiste Alayrac, Jeff Donahue, Pauline Luc, Antoine Miech, Iain Barr, Yana Hasson, Karel Lenc, Arthur Mensch, Katherine Millican, Malcolm Reynolds, Roman Ring, Eliza Rutherford, Serkan Cabi, Tengda Han, Zhitao Gong, Sina Samangooei, Marianne Monteiro, Jacob~L Menick, Sebastian Borgeaud, Andy Brock, Aida Nematzadeh, Sahand Sharifzadeh, Miko\l~aj Bi\'{n}kowski, Ricardo Barreira, Oriol Vinyals, Andrew Zisserman, and Kar\'{e}n Simonyan.
\newblock Flamingo: a visual language model for few-shot learning.
\newblock In \emph{NeurIPS}, volume~35, pp.\  23716--23736, 2022.

\bibitem[Anil et~al.(2023)Anil, Dai, Firat, Johnson, Lepikhin, Passos, Shakeri, Taropa, Bailey, Chen, et~al.]{PaLMv2_arXiv23}
Rohan Anil, Andrew~M Dai, Orhan Firat, Melvin Johnson, Dmitry Lepikhin, Alexandre Passos, Siamak Shakeri, Emanuel Taropa, Paige Bailey, Zhifeng Chen, et~al.
\newblock Palm 2 technical report.
\newblock \emph{arXiv preprint arXiv:2305.10403}, 2023.

\bibitem[Bai et~al.(2023{\natexlab{a}})Bai, Bai, Chu, Cui, Dang, Deng, Fan, Ge, Han, Huang, et~al.]{Qwen_arXiv23}
Jinze Bai, Shuai Bai, Yunfei Chu, Zeyu Cui, Kai Dang, Xiaodong Deng, Yang Fan, Wenbin Ge, Yu~Han, Fei Huang, et~al.
\newblock Qwen technical report.
\newblock \emph{arXiv preprint arXiv:2309.16609}, 2023{\natexlab{a}}.

\bibitem[Bai et~al.(2023{\natexlab{b}})Bai, Bai, Yang, Wang, Tan, Wang, Lin, Zhou, and Zhou]{QwenVL_arXiv23}
Jinze Bai, Shuai Bai, Shusheng Yang, Shijie Wang, Sinan Tan, Peng Wang, Junyang Lin, Chang Zhou, and Jingren Zhou.
\newblock Qwen-vl: A versatile vision-language model for understanding, localization, text reading, and beyond.
\newblock \emph{arXiv preprint arXiv:2308.12966}, 2023{\natexlab{b}}.

\bibitem[Bai et~al.(2025)Bai, Chen, Liu, Wang, Ge, Song, Dang, Wang, Wang, Tang, et~al.]{QwenVL2.5_arXiv25}
Shuai Bai, Keqin Chen, Xuejing Liu, Jialin Wang, Wenbin Ge, Sibo Song, Kai Dang, Peng Wang, Shijie Wang, Jun Tang, et~al.
\newblock Qwen2. 5-vl technical report.
\newblock \emph{arXiv preprint arXiv:2502.13923}, 2025.

\bibitem[Bi et~al.(2024)Bi, Chen, Chen, Chen, Dai, Deng, Ding, Dong, Du, Fu, et~al.]{Deepseek_arXiv24}
Xiao Bi, Deli Chen, Guanting Chen, Shanhuang Chen, Damai Dai, Chengqi Deng, Honghui Ding, Kai Dong, Qiushi Du, Zhe Fu, et~al.
\newblock Deepseek llm: Scaling open-source language models with longtermism.
\newblock \emph{arXiv preprint arXiv:2401.02954}, 2024.

\bibitem[Brown et~al.(2020)Brown, Mann, Ryder, Subbiah, Kaplan, Dhariwal, Neelakantan, Shyam, Sastry, Askell, et~al.]{GPT3_NeurIPS20}
Tom Brown, Benjamin Mann, Nick Ryder, Melanie Subbiah, Jared~D Kaplan, Prafulla Dhariwal, Arvind Neelakantan, Pranav Shyam, Girish Sastry, Amanda Askell, et~al.
\newblock Language models are few-shot learners.
\newblock In \emph{NeurIPS}, pp.\  1877--1901, 2020.

\bibitem[Caron et~al.(2021)Caron, Touvron, Misra, J{\'e}gou, Mairal, Bojanowski, and Joulin]{dino_ICCV21}
Mathilde Caron, Hugo Touvron, Ishan Misra, Herv{\'e} J{\'e}gou, Julien Mairal, Piotr Bojanowski, and Armand Joulin.
\newblock Emerging properties in self-supervised vision transformers.
\newblock In \emph{ICCV}, pp.\  9650--9660, 2021.

\bibitem[Chen et~al.(2024)Chen, Liu, Du, Pang, Liu, Sinha, Varakantham, and Lin]{DICE_arXiv24}
Changyu Chen, Zichen Liu, Chao Du, Tianyu Pang, Qian Liu, Arunesh Sinha, Pradeep Varakantham, and Min Lin.
\newblock Bootstrapping language models with dpo implicit rewards.
\newblock \emph{arXiv preprint arXiv:2406.09760}, 2024.

\bibitem[Chen et~al.(2025{\natexlab{a}})Chen, Chen, Wang, and Yang]{SEEDGRPO_arXiv25}
Minghan Chen, Guikun Chen, Wenguan Wang, and Yi~Yang.
\newblock Seed-grpo: Semantic entropy enhanced grpo for uncertainty-aware policy optimization.
\newblock \emph{arXiv preprint arXiv:2505.12346}, 2025{\natexlab{a}}.

\bibitem[Chen et~al.(2025{\natexlab{b}})Chen, Ge, Wang, Ge, Cheng, Shan, and Liu]{GRPOCARE_arXiv25}
Yi~Chen, Yuying Ge, Rui Wang, Yixiao Ge, Junhao Cheng, Ying Shan, and Xihui Liu.
\newblock Grpo-care: Consistency-aware reinforcement learning for multimodal reasoning.
\newblock \emph{arXiv preprint arXiv:2506.16141}, 2025{\natexlab{b}}.

\bibitem[Chowdhery et~al.(2022)Chowdhery, Narang, Devlin, Bosma, Mishra, Roberts, Barham, Chung, Sutton, Gehrmann, et~al.]{PaLM_arXiv22}
Aakanksha Chowdhery, Sharan Narang, Jacob Devlin, Maarten Bosma, Gaurav Mishra, Adam Roberts, Paul Barham, Hyung~Won Chung, Charles Sutton, Sebastian Gehrmann, et~al.
\newblock Palm: Scaling language modeling with pathways.
\newblock \emph{arXiv preprint arXiv:2204.02311}, 2022.

\bibitem[Chu et~al.(2025{\natexlab{a}})Chu, Zhai, Yang, Tong, Xie, Schuurmans, Le, Levine, and Ma]{SFTMemorizeRLGeneralize_arXiv25}
Tianzhe Chu, Yuexiang Zhai, Jihan Yang, Shengbang Tong, Saining Xie, Dale Schuurmans, Quoc~V Le, Sergey Levine, and Yi~Ma.
\newblock Sft memorizes, rl generalizes: A comparative study of foundation model post-training.
\newblock \emph{arXiv preprint arXiv:2501.17161}, 2025{\natexlab{a}}.

\bibitem[Chu et~al.(2025{\natexlab{b}})Chu, Huang, Zhang, Wei, and Wang]{GPG_arXiv25}
Xiangxiang Chu, Hailang Huang, Xiao Zhang, Fei Wei, and Yong Wang.
\newblock Gpg: A simple and strong reinforcement learning baseline for model reasoning.
\newblock \emph{arXiv preprint arXiv:2504.02546}, 2025{\natexlab{b}}.

\bibitem[Dai et~al.(2025{\natexlab{a}})Dai, Liu, and Si]{GRPOLambda_arXiv25}
Muzhi Dai, Shixuan Liu, and Qingyi Si.
\newblock Stable reinforcement learning for efficient reasoning.
\newblock \emph{arXiv preprint arXiv:2505.18086}, 2025{\natexlab{a}}.

\bibitem[Dai et~al.(2025{\natexlab{b}})Dai, Yang, and Si]{SGRPO_arXiv25}
Muzhi Dai, Chenxu Yang, and Qingyi Si.
\newblock S-grpo: Early exit via reinforcement learning in reasoning models.
\newblock \emph{arXiv preprint arXiv:2505.07686}, 2025{\natexlab{b}}.

\bibitem[Dai et~al.(2023)Dai, Li, Li, Tiong, Zhao, Wang, Li, Fung, and Hoi]{InstructBLIP_NeurIPS23}
Wenliang Dai, Junnan Li, Dongxu Li, Anthony Meng~Huat Tiong, Junqi Zhao, Weisheng Wang, Boyang Li, Pascale Fung, and Steven Hoi.
\newblock Instructblip: Towards general-purpose vision-language models with instruction tuning.
\newblock In \emph{NeurIPS}, 2023.

\bibitem[Ding et~al.(2025)Ding, Wang, Zeng, and Wang]{MGRPO_arXiv25}
Fei Ding, Baiqiao Wang, Zijian Zeng, and Youwei Wang.
\newblock Multi-layer grpo: Enhancing reasoning and self-correction in large language models.
\newblock \emph{arXiv preprint arXiv:2506.04746}, 2025.

\bibitem[Dosovitskiy et~al.(2021)Dosovitskiy, Beyer, Kolesnikov, Weissenborn, Zhai, Unterthiner, Dehghani, Minderer, Heigold, Gelly, et~al.]{ViT_ICLR21}
Alexey Dosovitskiy, Lucas Beyer, Alexander Kolesnikov, Dirk Weissenborn, Xiaohua Zhai, Thomas Unterthiner, Mostafa Dehghani, Matthias Minderer, Georg Heigold, Sylvain Gelly, et~al.
\newblock An image is worth 16x16 words: Transformers for image recognition at scale.
\newblock In \emph{ICLR}, 2021.

\bibitem[Fan et~al.(2025)Fan, Feng, Lyu, Zhou, and Yue]{SophiaVLR1_arXiv25}
Kaixuan Fan, Kaituo Feng, Haoming Lyu, Dongzhan Zhou, and Xiangyu Yue.
\newblock Sophiavl-r1: Reinforcing mllms reasoning with thinking reward.
\newblock \emph{arXiv preprint arXiv:2505.17018}, 2025.

\bibitem[Fang et~al.(2024)Fang, Zhu, Lu, Wang, Molchanov, Cho, Pavone, Han, and Yin]{VILA2_arXiv24}
Yunhao Fang, Ligeng Zhu, Yao Lu, Yan Wang, Pavlo Molchanov, Jang~Hyun Cho, Marco Pavone, Song Han, and Hongxu Yin.
\newblock $vila^2$: Vila augmented vila.
\newblock \emph{arXiv preprint arXiv:2407.17453}, 2024.

\bibitem[Guo et~al.(2025)Guo, Yang, Zhang, Song, Zhang, Xu, Zhu, Ma, Wang, Bi, et~al.]{DeepSeekR1_arXiv25}
Daya Guo, Dejian Yang, Haowei Zhang, Junxiao Song, Ruoyu Zhang, Runxin Xu, Qihao Zhu, Shirong Ma, Peiyi Wang, Xiao Bi, et~al.
\newblock Deepseek-r1: Incentivizing reasoning capability in llms via reinforcement learning.
\newblock \emph{arXiv preprint arXiv:2501.12948}, 2025.

\bibitem[Hu et~al.(2025)Hu, Zhang, Han, Jiang, Zhang, and Shum]{OpenReasoner_arXiv25}
Jingcheng Hu, Yinmin Zhang, Qi~Han, Daxin Jiang, Xiangyu Zhang, and Heung-Yeung Shum.
\newblock Open-reasoner-zero: An open source approach to scaling up reinforcement learning on the base model.
\newblock \emph{arXiv preprint arXiv:2503.24290}, 2025.

\bibitem[Huang et~al.(2025)Huang, Dai, Liu, He, Jiang, Song, Chen, Yao, and Song]{HintGRPO_ICCV25}
Qihan Huang, Weilong Dai, Jinlong Liu, Wanggui He, Hao Jiang, Mingli Song, Jingyuan Chen, Chang Yao, and Jie Song.
\newblock Boosting mllm reasoning with text-debiased hint-grpo.
\newblock In \emph{ICCV}, 2025.

\bibitem[Jaech et~al.(2024)Jaech, Kalai, Lerer, Richardson, El-Kishky, Low, Helyar, Madry, Beutel, Carney, et~al.]{OpenAIOone_arXiv24}
Aaron Jaech, Adam Kalai, Adam Lerer, Adam Richardson, Ahmed El-Kishky, Aiden Low, Alec Helyar, Aleksander Madry, Alex Beutel, Alex Carney, et~al.
\newblock Openai o1 system card.
\newblock \emph{arXiv preprint arXiv:2412.16720}, 2024.

\bibitem[Kingma \& Ba(2014)Kingma and Ba]{Adam_arXiv14}
Diederik~P Kingma and Jimmy Ba.
\newblock Adam: A method for stochastic optimization.
\newblock \emph{arXiv preprint arXiv:1412.6980}, 2014.

\bibitem[Li et~al.(2024)Li, Zhang, Zhang, Zhang, Li, Li, Ma, and Li]{LLaVANext_arXiv24}
Feng Li, Renrui Zhang, Hao Zhang, Yuanhan Zhang, Bo~Li, Wei Li, Zejun Ma, and Chunyuan Li.
\newblock Llava-next-interleave: Tackling multi-image, video, and 3d in large multimodal models.
\newblock \emph{arXiv preprint arXiv:2407.07895}, 2024.

\bibitem[Li et~al.(2022)Li, Li, Xiong, and Hoi]{BLIP_ICML22}
Junnan Li, Dongxu Li, Caiming Xiong, and Steven Hoi.
\newblock Blip: Bootstrapping language-image pre-training for unified vision-language understanding and generation.
\newblock In \emph{ICML}, pp.\  12888--12900. PMLR, 2022.

\bibitem[Li et~al.(2023)Li, Li, Savarese, and Hoi]{BLIPv2_ICML23}
Junnan Li, Dongxu Li, Silvio Savarese, and Steven Hoi.
\newblock Blip-2: Bootstrapping language-image pre-training with frozen image encoders and large language models.
\newblock In \emph{ICML}, pp.\  19730--19742. PMLR, 2023.

\bibitem[Li et~al.(2025)Li, Wei, Zheng, Huang, Kong, Sun, and Huang]{VisionMatters_arXiv25}
Yuting Li, Lai Wei, Kaipeng Zheng, Jingyuan Huang, Linghe Kong, Lichao Sun, and Weiran Huang.
\newblock Vision matters: Simple visual perturbations can boost multimodal math reasoning.
\newblock \emph{arXiv preprint arXiv:2506.09736}, 2025.

\bibitem[Lin et~al.(2023)Lin, Yin, Ping, Lu, Molchanov, Tao, Mao, Kautz, Shoeybi, and Han]{VILA_CVPR24}
Ji~Lin, Hongxu Yin, Wei Ping, Yao Lu, Pavlo Molchanov, Andrew Tao, Huizi Mao, Jan Kautz, Mohammad Shoeybi, and Song Han.
\newblock Vila: On pre-training for visual language models.
\newblock In \emph{CVPR}, 2023.

\bibitem[Liu et~al.(2023{\natexlab{a}})Liu, Li, Li, and Lee]{LLaVA15_CVPR24}
Haotian Liu, Chunyuan Li, Yuheng Li, and Yong~Jae Lee.
\newblock Improved baselines with visual instruction tuning.
\newblock In \emph{CVPR}, 2023{\natexlab{a}}.

\bibitem[Liu et~al.(2023{\natexlab{b}})Liu, Li, Wu, and Lee]{LLaVA_NeurIPS23}
Haotian Liu, Chunyuan Li, Qingyang Wu, and Yong~Jae Lee.
\newblock Visual instruction tuning.
\newblock In \emph{NeurIPS}, 2023{\natexlab{b}}.

\bibitem[Liu et~al.(2025{\natexlab{a}})Liu, Fang, Hu, Zhang, Zhou, Zhang, Tu, Lin, Huang, Song, et~al.]{DPOSurvey_arXiv25}
Shunyu Liu, Wenkai Fang, Zetian Hu, Junjie Zhang, Yang Zhou, Kongcheng Zhang, Rongcheng Tu, Ting-En Lin, Fei Huang, Mingli Song, et~al.
\newblock A survey of direct preference optimization.
\newblock \emph{arXiv preprint arXiv:2503.11701}, 2025{\natexlab{a}}.

\bibitem[Liu et~al.(2025{\natexlab{b}})Liu, Ni, Wu, Du, Dou, Wang, Pang, and Shieh]{NoisyRollout_arXiv25}
Xiangyan Liu, Jinjie Ni, Zijian Wu, Chao Du, Longxu Dou, Haonan Wang, Tianyu Pang, and Michael~Qizhe Shieh.
\newblock Noisyrollout: Reinforcing visual reasoning with data augmentation.
\newblock \emph{arXiv preprint arXiv:2504.13055}, 2025{\natexlab{b}}.

\bibitem[Liu et~al.(2025{\natexlab{c}})Liu, Qu, Zhong, Peng, Liu, Yu, and Jia]{VisionReasoner_arXiv25}
Yuqi Liu, Tianyuan Qu, Zhisheng Zhong, Bohao Peng, Shu Liu, Bei Yu, and Jiaya Jia.
\newblock Visionreasoner: Unified visual perception and reasoning via reinforcement learning.
\newblock \emph{arXiv preprint arXiv:2505.12081}, 2025{\natexlab{c}}.

\bibitem[Liu et~al.(2025{\natexlab{d}})Liu, Zhu, Shi, Zhang, Lou, Yang, Xi, Cao, Gu, Li, et~al.]{NVILA_CVPR25}
Zhijian Liu, Ligeng Zhu, Baifeng Shi, Zhuoyang Zhang, Yuming Lou, Shang Yang, Haocheng Xi, Shiyi Cao, Yuxian Gu, Dacheng Li, et~al.
\newblock Nvila: Efficient frontier visual language models.
\newblock In \emph{CVPR}, 2025{\natexlab{d}}.

\bibitem[Liu et~al.(2025{\natexlab{e}})Liu, Chen, Li, Qi, Pang, Du, Lee, and Lin]{DrGRPO_arXiv25}
Zichen Liu, Changyu Chen, Wenjun Li, Penghui Qi, Tianyu Pang, Chao Du, Wee~Sun Lee, and Min Lin.
\newblock Understanding r1-zero-like training: A critical perspective.
\newblock \emph{arXiv preprint arXiv:2503.20783}, 2025{\natexlab{e}}.

\bibitem[Liu et~al.(2025{\natexlab{f}})Liu, Sun, Zang, Dong, Cao, Duan, Lin, and Wang]{VRFT_arXiv25}
Ziyu Liu, Zeyi Sun, Yuhang Zang, Xiaoyi Dong, Yuhang Cao, Haodong Duan, Dahua Lin, and Jiaqi Wang.
\newblock Visual-rft: Visual reinforcement fine-tuning.
\newblock \emph{arXiv preprint arXiv:2503.01785}, 2025{\natexlab{f}}.

\bibitem[Lu et~al.(2021)Lu, Gong, Jiang, Qiu, Huang, Liang, and Zhu]{GeoThreeK_arXiv21}
Pan Lu, Ran Gong, Shibiao Jiang, Liang Qiu, Siyuan Huang, Xiaodan Liang, and Song-Chun Zhu.
\newblock Inter-gps: Interpretable geometry problem solving with formal language and symbolic reasoning.
\newblock \emph{arXiv preprint arXiv:2105.04165}, 2021.

\bibitem[Lu et~al.(2023)Lu, Bansal, Xia, Liu, Li, Hajishirzi, Cheng, Chang, Galley, and Gao]{MathVista_arXiv23}
Pan Lu, Hritik Bansal, Tony Xia, Jiacheng Liu, Chunyuan Li, Hannaneh Hajishirzi, Hao Cheng, Kai-Wei Chang, Michel Galley, and Jianfeng Gao.
\newblock Mathvista: Evaluating mathematical reasoning of foundation models in visual contexts.
\newblock \emph{arXiv preprint arXiv:2310.02255}, 2023.

\bibitem[Ma et~al.(2025)Ma, Liang, Qiang, Tang, Ma, Wong, Niu, Shen, He, Cui, et~al.]{ReLIFT_arXiv25}
Lu~Ma, Hao Liang, Meiyi Qiang, Lexiang Tang, Xiaochen Ma, Zhen~Hao Wong, Junbo Niu, Chengyu Shen, Runming He, Bin Cui, et~al.
\newblock Learning what reinforcement learning can't: Interleaved online fine-tuning for hardest questions.
\newblock \emph{arXiv preprint arXiv:2506.07527}, 2025.

\bibitem[OpenAI(2023)]{GPT4_arXiv23}
OpenAI.
\newblock Gpt-4 technical report.
\newblock \emph{arXiv preprint arXiv:2303.08774}, 2023.

\bibitem[Ouyang et~al.(2022)Ouyang, Wu, Jiang, Almeida, Wainwright, Mishkin, Zhang, Agarwal, Slama, Ray, et~al.]{HumanFeedback_NeurIPS22}
Long Ouyang, Jeffrey Wu, Xu~Jiang, Diogo Almeida, Carroll Wainwright, Pamela Mishkin, Chong Zhang, Sandhini Agarwal, Katarina Slama, Alex Ray, et~al.
\newblock Training language models to follow instructions with human feedback.
\newblock \emph{Advances in neural information processing systems}, 35:\penalty0 27730--27744, 2022.

\bibitem[Panda et~al.(2018)Panda, Zhang, Li, Lee, Lu, and Roy-Chowdhury]{WEBEmo_ECCV18}
Rameswar Panda, Jianming Zhang, Haoxiang Li, Joon-Young Lee, Xin Lu, and Amit~K Roy-Chowdhury.
\newblock Contemplating visual emotions: Understanding and overcoming dataset bias.
\newblock In \emph{ECCV}, pp.\  579--595, 2018.

\bibitem[Peng et~al.(2015)Peng, Chen, Sadovnik, and Gallagher]{EmotionSix_CVPR15}
Kuan-Chuan Peng, Tsuhan Chen, Amir Sadovnik, and Andrew~C Gallagher.
\newblock A mixed bag of emotions: Model, predict, and transfer emotion distributions.
\newblock In \emph{CVPR}, pp.\  860--868, 2015.

\bibitem[Pikus et~al.(2025)Pikus, Tiwari, and Ye]{HardExamplesGRPO_arXiv25}
Benjamin Pikus, Pratyush~Ranjan Tiwari, and Burton Ye.
\newblock Hard examples are all you need: Maximizing grpo post-training under annotation budgets.
\newblock \emph{arXiv preprint arXiv:2508.14094}, 2025.

\bibitem[Radford et~al.(2019)Radford, Wu, Child, Luan, Amodei, Sutskever, et~al.]{GPT2_OPENAI19}
Alec Radford, Jeffrey Wu, Rewon Child, David Luan, Dario Amodei, Ilya Sutskever, et~al.
\newblock Language models are unsupervised multitask learners.
\newblock \emph{OpenAI blog}, 1\penalty0 (8):\penalty0 9, 2019.

\bibitem[Radford et~al.(2021)Radford, Kim, Hallacy, Ramesh, Goh, Agarwal, Sastry, Askell, Mishkin, Clark, et~al.]{CLIP_ICML21}
Alec Radford, Jong~Wook Kim, Chris Hallacy, Aditya Ramesh, Gabriel Goh, Sandhini Agarwal, Girish Sastry, Amanda Askell, Pamela Mishkin, Jack Clark, et~al.
\newblock Learning transferable visual models from natural language supervision.
\newblock In \emph{ICML}, pp.\  8748--8763, 2021.

\bibitem[Rafailov et~al.(2023)Rafailov, Sharma, Mitchell, Manning, Ermon, and Finn]{DPO_NeurIPS23}
Rafael Rafailov, Archit Sharma, Eric Mitchell, Christopher~D Manning, Stefano Ermon, and Chelsea Finn.
\newblock Direct preference optimization: Your language model is secretly a reward model.
\newblock In \emph{NeurIPS}, pp.\  53728--53741, 2023.

\bibitem[Schulman et~al.(2017)Schulman, Wolski, Dhariwal, Radford, and Klimov]{PPO_arXiv17}
John Schulman, Filip Wolski, Prafulla Dhariwal, Alec Radford, and Oleg Klimov.
\newblock Proximal policy optimization algorithms.
\newblock \emph{arXiv preprint arXiv:1707.06347}, 2017.

\bibitem[Shao et~al.(2024)Shao, Wang, Zhu, Xu, Song, Bi, Zhang, Zhang, Li, Wu, et~al.]{Deepseekmath_arXiv24}
Zhihong Shao, Peiyi Wang, Qihao Zhu, Runxin Xu, Junxiao Song, Xiao Bi, Haowei Zhang, Mingchuan Zhang, YK~Li, Yang Wu, et~al.
\newblock Deepseekmath: Pushing the limits of mathematical reasoning in open language models.
\newblock \emph{arXiv preprint arXiv:2402.03300}, 2024.

\bibitem[Sheng et~al.(2025)Sheng, Zhang, Ye, Wu, Zhang, Zhang, Peng, Lin, and Wu]{Hybridflow_EuroSys25}
Guangming Sheng, Chi Zhang, Zilingfeng Ye, Xibin Wu, Wang Zhang, Ru~Zhang, Yanghua Peng, Haibin Lin, and Chuan Wu.
\newblock Hybridflow: A flexible and efficient rlhf framework.
\newblock In \emph{EuroSys}, pp.\  1279--1297, 2025.

\bibitem[Tang et~al.(2025)Tang, Dolga, Yoon, and Bogunovic]{wd1_arXiv25}
Xiaohang Tang, Rares Dolga, Sangwoong Yoon, and Ilija Bogunovic.
\newblock wd1: Weighted policy optimization for reasoning in diffusion language models.
\newblock \emph{arXiv preprint arXiv:2507.08838}, 2025.

\bibitem[Team et~al.(2025)Team, Du, Gao, Xing, Jiang, Chen, Li, Xiao, Du, Liao, et~al.]{Kimik1_arXiv25}
Kimi Team, Angang Du, Bofei Gao, Bowei Xing, Changjiu Jiang, Cheng Chen, Cheng Li, Chenjun Xiao, Chenzhuang Du, Chonghua Liao, et~al.
\newblock Kimi k1. 5: Scaling reinforcement learning with llms.
\newblock \emph{arXiv preprint arXiv:2501.12599}, 2025.

\bibitem[Touvron et~al.(2023)Touvron, Lavril, Izacard, Martinet, Lachaux, Lacroix, Rozi{\`e}re, Goyal, Hambro, Azhar, et~al.]{LLaMA_arXiv23}
Hugo Touvron, Thibaut Lavril, Gautier Izacard, Xavier Martinet, Marie-Anne Lachaux, Timoth{\'e}e Lacroix, Baptiste Rozi{\`e}re, Naman Goyal, Eric Hambro, Faisal Azhar, et~al.
\newblock Llama: Open and efficient foundation language models.
\newblock \emph{arXiv preprint arXiv:2302.13971}, 2023.

\bibitem[Wang et~al.(2025{\natexlab{a}})Wang, Ma, Reid, and Yaqub]{KRPO_arXiv25}
Hu~Wang, Congbo Ma, Ian Reid, and Mohammad Yaqub.
\newblock Kalman filter enhanced grpo for reinforcement learning-based language model reasoning.
\newblock \emph{arXiv preprint arXiv:2505.07527}, 2025{\natexlab{a}}.

\bibitem[Wang et~al.(2024)Wang, Pan, Shi, Lu, Ren, Zhou, Zhan, and Li]{MathVision_NeurIPS24}
Ke~Wang, Junting Pan, Weikang Shi, Zimu Lu, Houxing Ren, Aojun Zhou, Mingjie Zhan, and Hongsheng Li.
\newblock Measuring multimodal mathematical reasoning with math-vision dataset.
\newblock In \emph{NeurIPS}, pp.\  95095--95169, 2024.

\bibitem[Wang et~al.(2025{\natexlab{b}})Wang, Gao, Gu, Pu, Cui, Wei, Liu, Jing, Ye, Shao, et~al.]{Internvl35_arXiv25}
Weiyun Wang, Zhangwei Gao, Lixin Gu, Hengjun Pu, Long Cui, Xingguang Wei, Zhaoyang Liu, Linglin Jing, Shenglong Ye, Jie Shao, et~al.
\newblock Internvl3. 5: Advancing open-source multimodal models in versatility, reasoning, and efficiency.
\newblock \emph{arXiv preprint arXiv:2508.18265}, 2025{\natexlab{b}}.

\bibitem[Wang et~al.(2023)Wang, Bao, Dong, Bjorck, Peng, Liu, Aggarwal, Mohammed, Singhal, Som, and Wei]{BEiT3_CVPR23}
Wenhui Wang, Hangbo Bao, Li~Dong, Johan Bjorck, Zhiliang Peng, Qiang Liu, Kriti Aggarwal, Owais~Khan Mohammed, Saksham Singhal, Subhojit Som, and Furu Wei.
\newblock Image as a foreign language: {BEiT} pretraining for vision and vision-language tasks.
\newblock In \emph{CVPR}, 2023.

\bibitem[Wen et~al.(2025)Wen, Cai, Xiao, He, An, Duan, Du, Liu, Tang, Lv, et~al.]{LightROne_arXiv25}
Liang Wen, Yunke Cai, Fenrui Xiao, Xin He, Qi~An, Zhenyu Duan, Yimin Du, Junchen Liu, Lifu Tang, Xiaowei Lv, et~al.
\newblock Light-r1: Curriculum sft, dpo and rl for long cot from scratch and beyond.
\newblock \emph{arXiv preprint arXiv:2503.10460}, 2025.

\bibitem[Xiong et~al.(2025)Xiong, Yao, Xu, Pang, Wang, Sahoo, Li, Jiang, Zhang, Xiong, et~al.]{ReinforceRej_arXiv25}
Wei Xiong, Jiarui Yao, Yuhui Xu, Bo~Pang, Lei Wang, Doyen Sahoo, Junnan Li, Nan Jiang, Tong Zhang, Caiming Xiong, et~al.
\newblock A minimalist approach to llm reasoning: from rejection sampling to reinforce.
\newblock \emph{arXiv preprint arXiv:2504.11343}, 2025.

\bibitem[Xu \& Ding(2025)Xu and Ding]{SPO_arXiv25}
Zhongwen Xu and Zihan Ding.
\newblock Single-stream policy optimization.
\newblock \emph{arXiv preprint arXiv:2509.13232}, 2025.

\bibitem[Yang et~al.(2025{\natexlab{a}})Yang, Li, Yang, Zhang, Hui, Zheng, Yu, Gao, Huang, Lv, et~al.]{Qwen3_arXiv25}
An~Yang, Anfeng Li, Baosong Yang, Beichen Zhang, Binyuan Hui, Bo~Zheng, Bowen Yu, Chang Gao, Chengen Huang, Chenxu Lv, et~al.
\newblock Qwen3 technical report.
\newblock \emph{arXiv preprint arXiv:2505.09388}, 2025{\natexlab{a}}.

\bibitem[Yang et~al.(2025{\natexlab{b}})Yang, Yu, Li, Liu, Huang, Huang, Jiang, Tu, Zhang, Zhou, et~al.]{Qwen2.5_arXiv25}
An~Yang, Bowen Yu, Chengyuan Li, Dayiheng Liu, Fei Huang, Haoyan Huang, Jiandong Jiang, Jianhong Tu, Jianwei Zhang, Jingren Zhou, et~al.
\newblock Qwen2. 5-1m technical report.
\newblock \emph{arXiv preprint arXiv:2501.15383}, 2025{\natexlab{b}}.

\bibitem[Yang et~al.(2023)Yang, Huang, Ding, Lischinski, Cohen-Or, and Huang]{EmoSet_ICCV23}
Jingyuan Yang, Qirui Huang, Tingting Ding, Dani Lischinski, Danny Cohen-Or, and Hui Huang.
\newblock Emoset: A large-scale visual emotion dataset with rich attributes.
\newblock In \emph{ICCV}, pp.\  20383--20394, 2023.

\bibitem[Yang et~al.(2025{\natexlab{c}})Yang, Guo, Huang, Liang, Wang, and Tang]{TreeRPO_arXiv25}
Zhicheng Yang, Zhijiang Guo, Yinya Huang, Xiaodan Liang, Yiwei Wang, and Jing Tang.
\newblock Treerpo: Tree relative policy optimization.
\newblock \emph{arXiv preprint arXiv:2506.05183}, 2025{\natexlab{c}}.

\bibitem[Yao et~al.(2025)Yao, Yin, Zhang, Yang, Wang, Wu, Su, Shen, Qiu, Tao, et~al.]{ShareGRPO_ICCV25}
Huanjin Yao, Qixiang Yin, Jingyi Zhang, Min Yang, Yibo Wang, Wenhao Wu, Fei Su, Li~Shen, Minghui Qiu, Dacheng Tao, et~al.
\newblock R1-sharevl: Incentivizing reasoning capability of multimodal large language models via share-grpo.
\newblock In \emph{ICCV}, 2025.

\bibitem[Yu et~al.(2025)Yu, Zhang, Zhu, Yuan, Zuo, Yue, Fan, Liu, Liu, Liu, et~al.]{DAPO_arXiv25}
Qiying Yu, Zheng Zhang, Ruofei Zhu, Yufeng Yuan, Xiaochen Zuo, Yu~Yue, Tiantian Fan, Gaohong Liu, Lingjun Liu, Xin Liu, et~al.
\newblock Dapo: An open-source llm reinforcement learning system at scale.
\newblock \emph{arXiv preprint arXiv:2503.14476}, 2025.

\bibitem[Yue et~al.(2025)Yue, Chen, Lu, Zhao, Wang, Song, and Huang]{DoesRLincentiveReasoning_arXiv25}
Yang Yue, Zhiqi Chen, Rui Lu, Andrew Zhao, Zhaokai Wang, Shiji Song, and Gao Huang.
\newblock Does reinforcement learning really incentivize reasoning capacity in llms beyond the base model?
\newblock \emph{arXiv preprint arXiv:2504.13837}, 2025.

\bibitem[Zhang et~al.(2024{\natexlab{a}})Zhang, Shen, Zhao, Zhang, Xu, Dou, and Bian]{PFPPO_arXiv24}
Chuheng Zhang, Wei Shen, Li~Zhao, Xuyun Zhang, Xiaolong Xu, Wanchun Dou, and Jiang Bian.
\newblock Policy filtration for rlhf to mitigate noise in reward models.
\newblock \emph{arXiv preprint arXiv:2409.06957}, 2024{\natexlab{a}}.

\bibitem[Zhang et~al.(2025)Zhang, Huang, Yao, Liu, Zhang, Lu, and Tao]{StepGRPO_arXiv25}
Jingyi Zhang, Jiaxing Huang, Huanjin Yao, Shunyu Liu, Xikun Zhang, Shijian Lu, and Dacheng Tao.
\newblock R1-vl: Learning to reason with multimodal large language models via step-wise group relative policy optimization.
\newblock \emph{arXiv preprint arXiv:2503.12937}, 2025.

\bibitem[Zhang \& Zuo(2025)Zhang and Zuo]{GRPOLEAD_arXiv25}
Jixiao Zhang and Chunsheng Zuo.
\newblock Grpo-lead: A difficulty-aware reinforcement learning approach for concise mathematical reasoning in language models.
\newblock \emph{arXiv preprint arXiv:2504.09696}, 2025.

\bibitem[Zhang et~al.(2024{\natexlab{b}})Zhang, Jiang, Zhang, Lin, Guo, Qiu, Zhou, Lu, Chang, Qiao, et~al.]{MathVerse_ECCV24}
Renrui Zhang, Dongzhi Jiang, Yichi Zhang, Haokun Lin, Ziyu Guo, Pengshuo Qiu, Aojun Zhou, Pan Lu, Kai-Wei Chang, Yu~Qiao, et~al.
\newblock Mathverse: Does your multi-modal llm truly see the diagrams in visual math problems?
\newblock In \emph{ECCV}, pp.\  169--186. Springer, 2024{\natexlab{b}}.

\bibitem[Zheng et~al.(2025)Zheng, Lu, Wang, Feng, Kuang, and Xiong]{EasyROne_2025}
Yaowei Zheng, Junting Lu, Shenzhi Wang, Zhangchi Feng, Dongdong Kuang, and Yuwen Xiong.
\newblock Easyr1: An efficient, scalable, multi-modality rl training framework, 2025.

\bibitem[Zhu et~al.(2024)Zhu, Zhu, Liu, Ou, Mou, and Tang]{LLaVAPhi_arXiv24}
Yichen Zhu, Minjie Zhu, Ning Liu, Zhicai Ou, Xiaofeng Mou, and Jian Tang.
\newblock Llava-phi: Efficient multi-modal assistant with small language model.
\newblock \emph{arXiv preprint arXiv:2401.02330}, 2024.

\end{thebibliography}
\bibliographystyle{iclr2026_conference}

\appendix

\clearpage
\section*{Appendix}

\section{Notation and Algorithm}

We provide the notation table in \cref{tab:notation} and proposed method algorithm in \cref{alg:ours}.

\begin{algorithm}[h]
\caption{\oursabbrv{}}
\label{alg:ours}
\SetAlgoLined
\SetNoFillComment
\small{\KwIn{Reference model $\pi_{ref}$, old model $\pi_{old}$, current policy model $\pi_\theta$, group size $G$, Training Step $E$, Current Step $e$, Training Batch $B$, Question Distribution $\rho_Q$, Query Prompt $q$ }}



Initialize $\pi_\theta \leftarrow \pi_{ref}$ 

\For {$e=1, 2, ..., E$}{

$\pi_{old} \leftarrow \pi_\theta$ 

$q  \sim \rho_Q$, $o=\{o_i\}_{i=1}^G  \sim \pi_{old}(\cdot|q)$ {\footnotesize{\color{DarkBlue}{\tcp*{Sample prompt with $G$ trajectory.}}}}

$\bm{r} = \{r_i\}_{i=1}^G = R(o)  $ {\footnotesize{\color{DarkBlue}{\tcp*{Measure trajectory reward.}}}}

$\mu = \frac{1}{G} \sum_{r_i \in \bm{r}} r_i$,  $\sigma = \sqrt{\frac{1}{G} \sum_{i=1}^{G} \left(r_i - \mu \right)^2}$ {\footnotesize{\color{DarkBlue}{\tcp*{Calculate static information.}}}}

%
\emph{\firstmodule}\;

$\hat{A}_i \leftarrow (\bm{r}, \mu, \sigma)$ via \cref{eq:originaladvantage}, $\hat{A}_i^{\firstmdouleabbrv} \leftarrow (\bm{r}, \mu)$ via \cref{eq:proposedadvantage} {\footnotesize{\color{DarkBlue}{\tcp*{Measure advantage.}}}}

\BlankLine

\emph{\secondmodule}\;

$N =  \sum_{i=1}^{G} \mathbf{1}_{\{r_i = 1\}}$, $p=\frac{N}{G}$
{\footnotesize{\color{DarkBlue}{\tcp*{Calculate trajectory success ratio.}}}}

$\lambda(p)= 1 - 4p(1-p)$ {\footnotesize{\color{DarkBlue}{\tcp*{Measure trajectory maturity degree.}}}}

\BlankLine

$\hat{A}_i^* \leftarrow (\hat{A}_i, \hat{A}_i^{\firstmdouleabbrv}, \lambda)$ through \cref{eq:mixadvantage} {\footnotesize{\color{DarkBlue}{\tcp*{Mixed advantage.}}}}

$\mathcal{J}_{GRPO}(\theta) \leftarrow (\hat{A}_i^*, o, \pi_\theta, \pi_{old}, \pi_{ref})$, $\theta = \theta - \eta \nabla \mathcal{J}_{GRPO}(\theta)$  {\footnotesize{\color{DarkBlue}{\tcp*{Update Weight.}}}}

}
\end{algorithm}

\begin{table*}[ht]\small
\captionsetup{font=small}
\caption{\textbf{Notation used in MAPO}. Summary of key variables and operations in our method. The \textit{Definition} column indicates where each symbol first appears in the main text.}
\label{tab:notation}
\vspace{-10pt}
\centering
\resizebox{\textwidth}{!}{
\setlength\tabcolsep{20pt} 
\renewcommand\arraystretch{1}
\begin{tabular}{c||c|c}
\hline\thickhline
\rowcolor{gray!20}
\textbf{Symbol} & \textbf{Description} & \textbf{Definition} \\
\hline\hline
\rowcolor{rxklightblue}
$q$ & Query prompt sampled from distribution $\rho_Q$ & \cref{eq:optimize} \\
$o_i$ & $i$-th trajectory (rollout) sampled from $\pi_{\text{old}}$ & \cref{eq:optimize} \\
\rowcolor{rxklightblue}
$r_i$ & Reward assigned to trajectory $o_i$ & \cref{eq:optimize} \\
$G$ & Group size (number of rollouts per query) & \cref{eq:optimize} \\
\rowcolor{rxklightblue}
$\pi_\theta, \pi_{\text{old}}, \pi_{\text{ref}}$ & Current, old, and reference policy models & \cref{eq:optimize} \\
$J_{\text{GRPO}}(\theta)$ & Group Relative Policy Optimization objective & \cref{eq:optimize} \\
\rowcolor{rxklightblue}
$\beta$ & KL regularization coefficient & \cref{eq:optimize} \\
$f_\epsilon(x,y)$ & Clipping function $\min(xy, \mathrm{clip}(x,1-\epsilon,1+\epsilon)y)$ & \cref{eq:optimize} \\
\rowcolor{rxklightblue}
$\mathbb{D}_{KL}[\pi_\theta \| \pi_{\text{ref}}]$ & KL divergence between policy and reference model & \cref{eq:optimize} \\
$R(q,o_i)$ & Reward function & \cref{sec:preliminary} \\
\rowcolor{rxklightblue}
$\mu, \sigma$ & Mean and standard deviation of rewards in group & \cref{eq:originaladvantage} \\
$\hat{A}_i$ & Standardized advantage $\tfrac{r_i-\mu}{\sigma}$ & \cref{eq:originaladvantage} \\
\rowcolor{rxklightblue}
$N$ & Number of successful trajectories in a group & \cref{successful_trajectory} \\
$p$ & Empirical success ratio $p=\tfrac{N}{G}$ & \cref{successful_trajectory} \\
\rowcolor{rxklightblue}
$\hat{A}_i^{\text{APD}}$ & Advantage Percent Deviation $\tfrac{r_i-\mu}{\mu+\epsilon}$ & \cref{eq:proposedadvantage} \\
$\lambda(p)$ & Trajectory maturity degree $1-4p(1-p)$ & \cref{eq:trajeorycertainty} \\
\rowcolor{rxklightblue}
$\hat{A}_i^*$ & Mixed advantage combining $\hat{A}_i$ and $\hat{A}_i^{\text{APD}}$ & \cref{eq:mixadvantage} \\
$r_{\text{Format}}, r_{\text{Accuracy}}$ & Format reward and accuracy reward & \cref{fig:advantagediscussion} \\
\rowcolor{rxklightblue}
$\varrho(p)$ & Gradient ratio $\nabla_\theta J_{\text{MAPO}} / \nabla_\theta J_{\text{GRPO}}$ & \cref{eq:ratio} \\
$\mathcal{S}$ & Training distribution & \cref{sec:Setup} \\
\rowcolor{rxklightblue}
$\mathcal{T}=\{T_t\}_{t=1}^{|\mathcal{T}|}$ & Set of unseen test datasets & \cref{sec:Setup} \\
$|\mathcal{T}|$ & Number of unseen test datasets & \cref{sec:Setup} \\
\rowcolor{rxklightblue}
$\mathcal{A}^S, \mathcal{A}^T, \bar{\mathcal{A}}$ 
& In-domain, out-of-domain and average accuracy
& \cref{sec:Setup} \\
\hline
\thickhline
\end{tabular}}
\vspace{-10pt}
\end{table*}

\section{Experimental Informataion}

\subsection{Dataset Introduction}
\label{sec:datasetintrouction}
We use the following two datasets from the mathematics and emotional tasks for experiments. 
\begin{fullitemize}
    \item \raisebox{-0.2\height}{\includegraphics[width=0.023\textwidth]{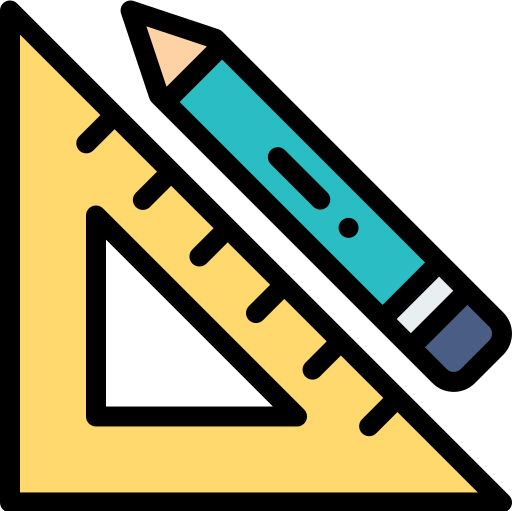}} \geothreek{} \pub{arXiv'21} {\cite{GeoThreeK_arXiv21}} Designed for geometry problem solving, this dataset contains images, text, and formulas that require models to perform joint visual–symbolic reasoning.

    \item \raisebox{-0.2\height}{\includegraphics[width=0.023\textwidth]{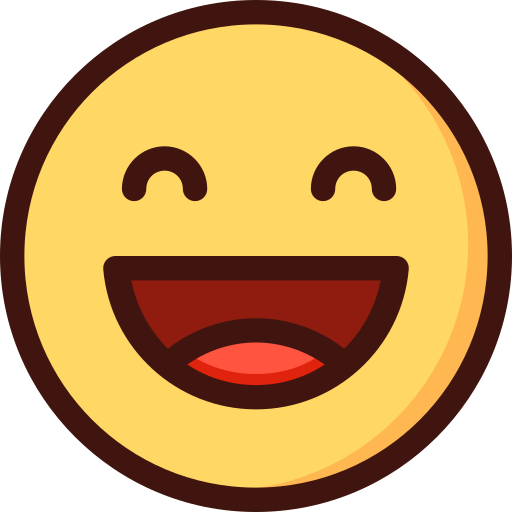}} \emoset{} \pub{ICCV'23} {\cite{EmoSet_ICCV23}} This large-scale collection targets visual emotion recognition, covering diverse scenes and a broad range of emotion categories.
\end{fullitemize}
Furthermore, for above two scenarios, we respectively conduct the evaluation on the following out-of-domain datasets to vaildate its generalization ability.

\begin{fullitemize}
\item \raisebox{-0.2\height}{\includegraphics[width=0.023\textwidth]{Figure/math.png}} \mathvista{} \pub{arXiv'23} {\cite{MathVista_arXiv23}} Serving as a benchmark for visual mathematical reasoning, spanning algebra, geometry, and word problems for cross-domain generalization.

\item \raisebox{-0.2\height}{\includegraphics[width=0.023\textwidth]{Figure/math.png}} \mathvision{} \pub{NeurIPS'24} {\cite{MathVision_NeurIPS24}} Proposed for multimodal mathematical reasoning, it emphasizes inference across visual diagrams and natural language expressions.

\item \raisebox{-0.2\height}{\includegraphics[width=0.023\textwidth]{Figure/math.png}} \mathverse{} \pub{ECCV'24} {\cite{MathVerse_ECCV24}} Built to assess model understanding of complex charts, geometric figures, and formula-rich inputs, emphasizing visual interpretation in reasoning.

\item \raisebox{-0.2\height}{\includegraphics[width=0.023\textwidth]{Figure/emo.png}} \webemo{} \pub{ECCV'18} {\cite{WEBEmo_ECCV18}} Comprising millions of web images, this dataset spans 7 high-level emotion categories and supports recognition and cross-domain emotion analysis.

\item \raisebox{-0.2\height}{\includegraphics[width=0.023\textwidth]{Figure/emo.png}} \emotionsix{} \pub{CVPR'15} \cite{EmotionSix_CVPR15} A classic benchmark for visual emotion recognition, consisting of 6 basic emotion categories and widely used for standard evaluation.
\end{fullitemize}

\begin{figure*}[h]
\centering
\begin{center}
\includegraphics[width=0.9\textwidth]{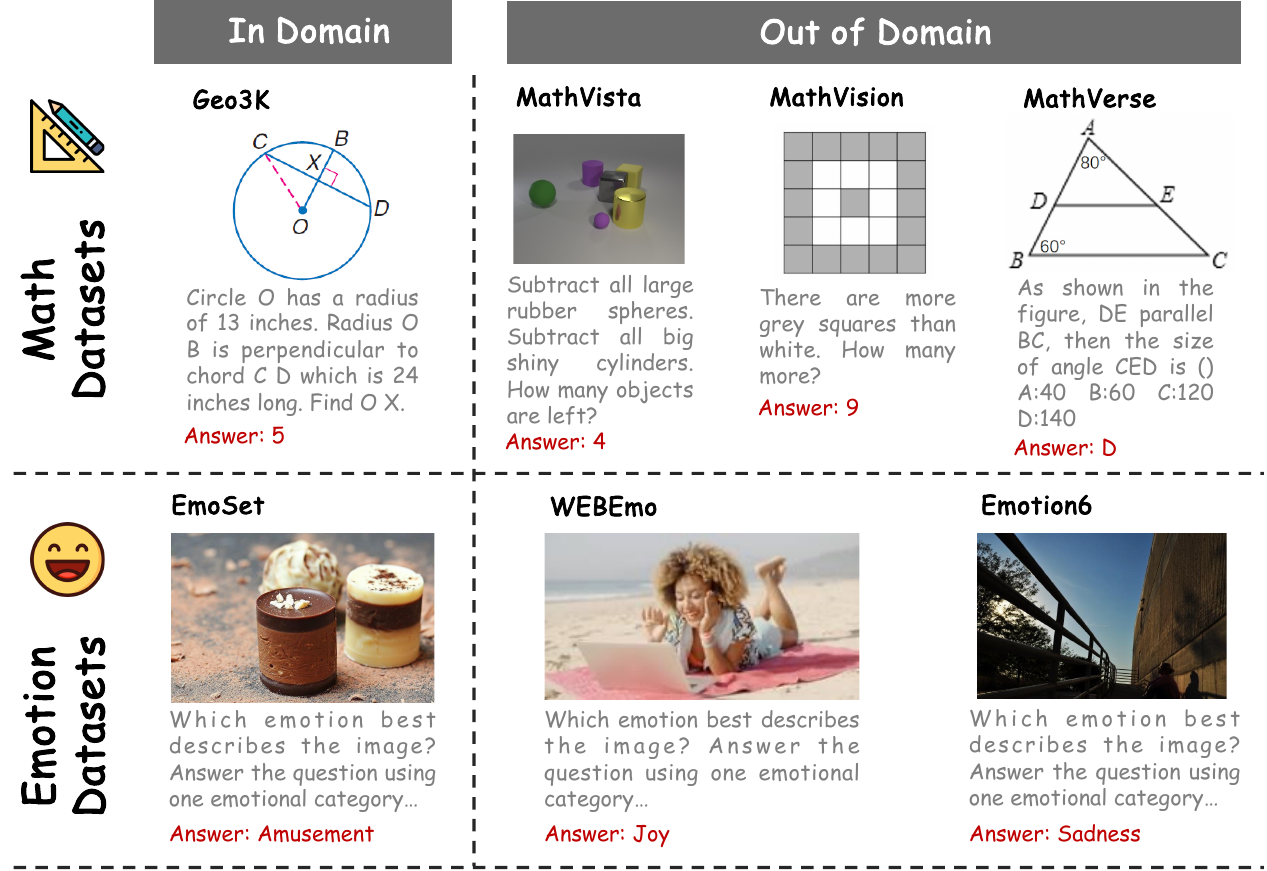}
\end{center}
\vspace{-10pt}
\captionsetup{font=small}
\caption{\textbf{Data Visualization} from the In Domain and Out of Domain datasets used in our experiments.}
\label{fig:case}
\vspace{-10pt}
\end{figure*}

We plot a detailed dataset case illustration in \cref{fig:case}.

\subsection{Implementation Details}
\label{sec:trainingdetails}
We conduct the experiments on the EasyR1\footnote{https://github.com/hiyouga/EasyR1}{\cite{EasyROne_2025}} as our reinforcement learning training framework, which is built on Verl\footnote{https://github.com/volcengine/verl} {\cite{Hybridflow_EuroSys25}}. The rollout batch size is set to $512$, and the global batch size is $128$. The rollout temperature during training is fixed at $1.0$, with Top-$p$ set to $0.99$. To mitigate token-length bias, we compute the policy loss using a token-mean aggregation strategy. The vision tower of Qwen2.5-VL-7B is fine-tuned without freezing. The optimizer is AdamW {\cite{Adam_arXiv14}} with a learning rate of $1 \times 10^{-6}$, and the KL coefficient $\beta$ is set to $1 \times 10^{-2}$. For the validation setting, we set the temperature to $0.5$. The maximum number of tokens to generate is $2048$, and Top-$p$ sampling is $0.95$. The training epoch is respectively set as $T=20$ and $T=15$ for \geothreek{} and \emoset{}. Consequently, the training step is $E=80$ for \geothreek{} and $E=60$ \emoset{}. These configurations are consistent with the EasyR1.

\subsection{Visualization Analysis}
\label{sec:visualization}

\begin{figure*}[t]
\centering
\begin{center}
\includegraphics[width=\linewidth]{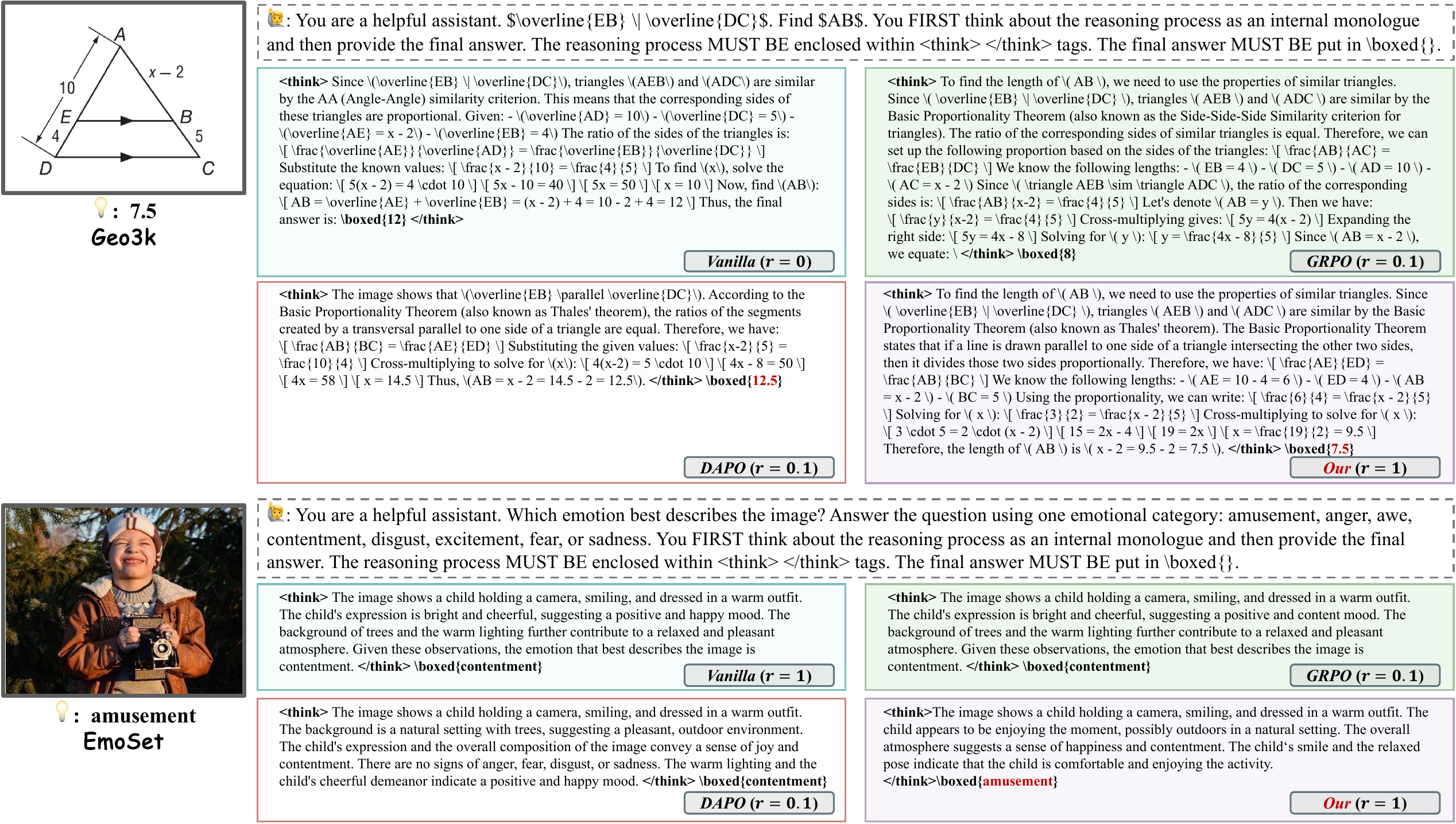}
\end{center}
\vspace{-10pt}
\captionsetup{font=small}
\caption{\textbf{In-Domain Case visualization}.} 
\label{fig:indomain_visualization}
\vspace{-10pt}
\end{figure*}

\begin{figure*}[t]
\centering
\begin{center}
\includegraphics[width=\linewidth]{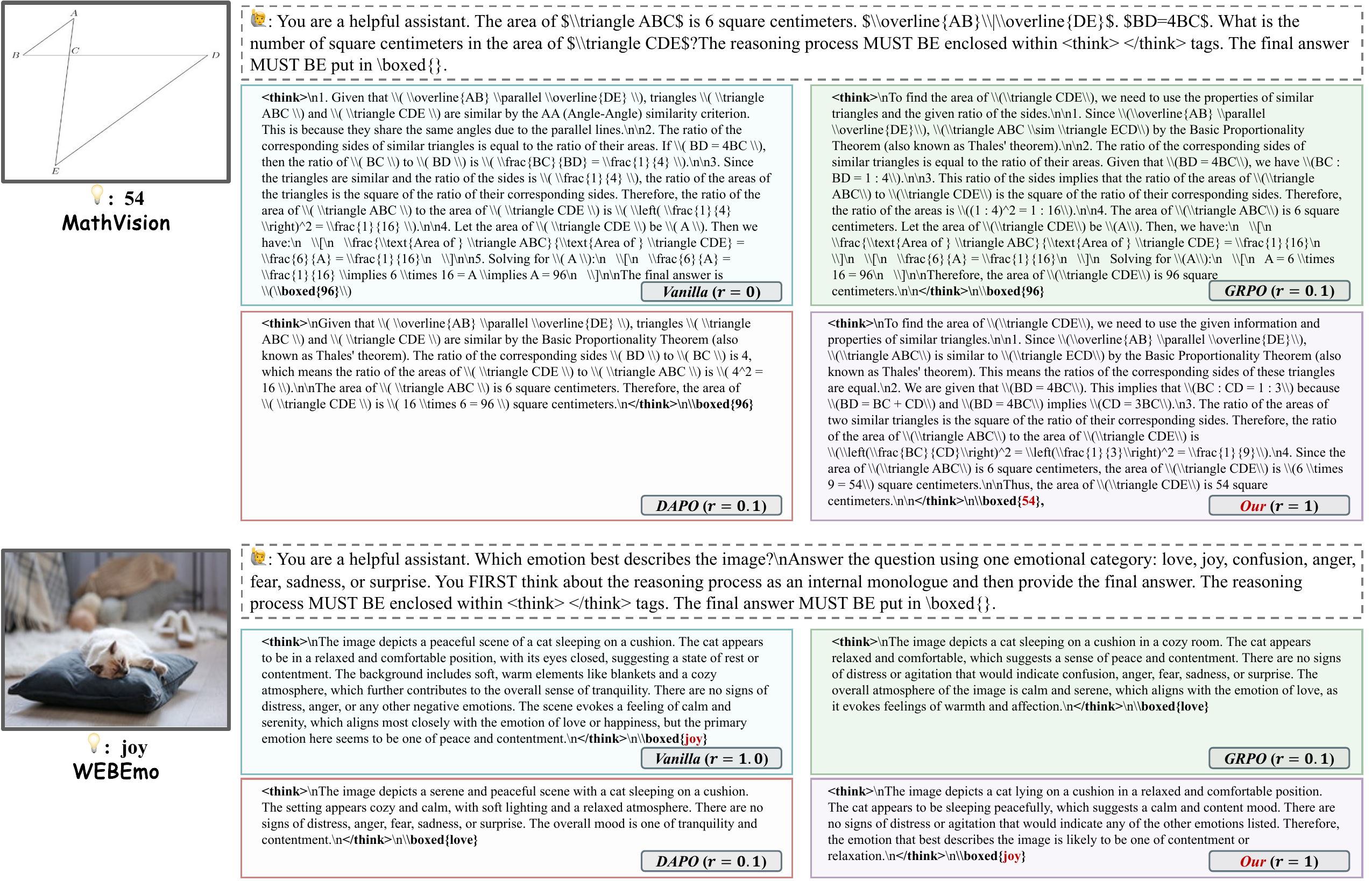}
\end{center}
\vspace{-10pt}
\captionsetup{font=small}
\caption{\textbf{Out-of-Domain Case Visualization}.} 
\label{fig:ooddomain_visualization}
\vspace{-10pt}
\end{figure*}

We present the output cases for both in-domain (\cref{fig:indomain_visualization}) and out-of-domain distributions (\cref{fig:ooddomain_visualization}). For samples with high certainty, the existing \grpoabbrv{} leads to abnormal behavior. Both \grpoabbrv{} and \dapo{} exhibit degradation on the in-domain dataset \emoset{} and the out-of-domain dataset \webemo{} in cases where the vanilla \qwentwofivevlsevenins version could correctly answer. This suggests that the existing advantage distorts the optimization direction for samples with high trajectory certainty, ultimately leading to a performance decrease.

\section{Theoretical Analysis}
\label{sec:theoryanalysis}
To better understand how \oursabbrv{} reshapes the optimization dynamics compared with \grpoabbrv{}, we provide a gradient-level analysis. 
\textit{Without loss of generality}, we simplify the gradient analysis by ignoring clipping and KL regularization and modeling the reward as accuracy, i.e., a Bernoulli variable.
For a prompt with $G$ rollouts and Bernoulli rewards $R_k\!\in\!\{0,1\}$, define $A_k=R_k-\mu$ and let $p=\frac{N}{G}$, where $N = \sum_{i=1}^{G} \mathbf{1}_{\{R_i = 1\}}$. Then
$\mu=\frac{1}{G}\sum_k R_k=p$ and $\sigma=\sqrt{p(1-p)}$.
Ignoring clipping and KL, the gradient of the objective is
\[
\nabla_\theta \mathcal{J}
=\mathbb{E}\!\Big[\sum_{k,t} r_{k,t}\,\hat A_k\,\nabla_\theta\log\pi_\theta(a_{k,t}\!\mid s_{k,t})\Big],
\quad r_{k,t}=\frac{\pi_\theta(a_{k,t}\mid s_{k,t})}{\pi_{\text{old}}(a_{k,t}\mid s_{k,t})}.
\]

For GRPO, the advantage is $\hat A_k^{\mathrm G}=A_k/\sigma$. 

For MAPO, 
\[
\hat A_k^{\mathrm M}=(1-\lambda(p))\frac{A_k}{\sigma}
+\lambda(p)\frac{A_k}{p},\qquad \lambda(p)=1-4p(1-p).
\]
Hence, for any trajectory, we define the ratio of the gradient as:
\[
\boxed{\ \varrho(p)\triangleq \frac{\nabla_\theta \mathcal J_{\mathrm{MAPO}}}{\nabla_\theta \mathcal J_{\mathrm{GRPO}}}
=\frac{\hat A_k^{\mathrm M}}{\hat A_k^{\mathrm G}}
=(1-\lambda(p))+\lambda(p)\frac{\sigma}{p}
=(1-\lambda(p))+\lambda(p)h(p)\, }
\]
where $h(p)=\sqrt{\frac{1-p}{p}}$.

Next, we analyze the property of $\varrho(p)$.
Since $h$ is smooth on $(0,1)$ with
\[
h'(p)= -\frac{1}{2p^2\,h(p)}
= -\frac{1}{2\,p^{3/2}\sqrt{1-p}}<0,
\]
the derivative of $\varrho$ is
\[
\varrho'(p)=4(1-2p)\big(1-h(p)\big)
+\big(1-4p(1-p)\big)\,h'(p).
\]
For $p \in (0,1)$, we obtain that $\varrho'(p) \leq 0$, and $\varrho(\tfrac12) = 1$. Thus, we have:
\[
\begin{cases}
\varrho(p) > 1, & p\in(0,\tfrac12),\\[4pt]
\varrho(p) = 1, & p = \tfrac{1}{2},\\[4pt]
0 < \varrho(p) < 1, & p\in(\tfrac12,1).
\end{cases}
\]
Which implies that \oursabbrv{} leads to amplified gradients than \grpoabbrv{} on harder samples (with $p<\tfrac12$), and smaller updates on easier samples (with $p>\tfrac12$). This leads to the conclusion in \cref{sec:discussionandlimitation}.

\noindent \textbf{Acknowledgement}.
I sincerely thank Haoyu Wang (ByteDance) and Jiamei Yan (ByteDance) for their valuable collaboration and support. This work would not have been possible without them. 
\end{document}